\documentclass{IEEEtran}

\usepackage{bm}
\usepackage{amssymb}
\usepackage{nicefrac} 
\usepackage{graphicx}
\usepackage{booktabs}
\usepackage{amsfonts}
\usepackage{amsmath}
\usepackage{amssymb}
\usepackage{pifont}
\usepackage{nicefrac} 
\usepackage{algorithm}
\usepackage{algorithmicx}
\usepackage{algpseudocode}
\usepackage{dsfont}
\usepackage{authblk}
\usepackage{bbm}
\usepackage{graphicx}
\usepackage{epstopdf}
\usepackage{subfigure}
\usepackage{stfloats}
\usepackage{eqnarray}
\usepackage{makecell}
\usepackage{multirow}
\usepackage{enumerate}
\usepackage{booktabs}
\usepackage{cite}
\usepackage{balance}
\usepackage{color}
\usepackage{bm}
\usepackage{caption}
\usepackage{mathtools}
\usepackage{url}

\renewcommand{\IEEEQED}{\IEEEQEDopen}

\newcommand{\oo}[1]{{\cal O}(#1)}

\newcommand{\inp}[1]{\langle#1\rangle}

\newcommand{\pp}[1]{\mathds{P}(#1)}
\newcommand{\ee}[1]{\mathds{E}[#1]}
\newcommand{\EE}[1]{\mathds{E}\Big[#1\Big]}

\newcommand{\norm}[1]{\|#1\|}
\newcommand{\normtv}[1]{\|#1\|_{\mbox{\tiny TV}}}
\newcommand{\NORM}[1]{\Big\|#1\Big\|}
\newcommand{\NORMTV}[1]{\Big\|#1\Big\|_{\mbox{\tiny TV}}}

\DeclareMathOperator{\proj}{\mbox{proj}}

\newcommand{\tar}{\mbox{\tiny tar}}

\newtheorem{theorem}{\textbf{Theorem}}

\newtheorem{lemma}{\textbf{Lemma}}

\newtheorem{remark}{\textbf{Remark}}

\newtheorem{corollary}{\textbf{Corollary}}
\newtheorem{assump}{\textbf{Assumption}}

\usepackage{balance}
\usepackage{textcomp}
\begin{document}
\title{Heavy-Ball Momentum Accelerated Actor-Critic With Function Approximation}
\author{Yanjie Dong, Haijun Zhang,~\IEEEmembership{Fellow, IEEE}, Gang Wang,~\IEEEmembership{Senior Member, IEEE}, Shisheng Cui, and Xiping Hu
\thanks{This work was supported by the National Natural Science Foundation of China under Grants 62102266, U23B2059, 62173034 and the Pearl River Talent Recruitment Program of Guangdong Province under Grant 2019ZT08X603.}
	\thanks{Y. Dong and X. Hu are with the Artificial Intelligence Research Institute and Guangdong-Hong Kong-Macao Joint Laboratory for Emotional Intelligence and Pervasive Computing, Shenzhen MSU-BIT University, Shenzhen 518172, China.}
	\thanks{H. Zhang is with the Beijing Engineering and Technology Research Center for Convergence Networks and Ubiquitous Services, University of Science and Technology Beijing, Beijing, China.}
%	\thanks{K. Xing and X. Hu are with Artificial Intelligence Research Institute and Guangdong-Hong Kong-Macao Joint Laboratory for Emotional Intelligence and Pervasive Computing, Shenzhen MSU-BIT University, Shenzhen 518172, China, and also with the School of Medical Technology, Beijing Institute of Technology, Beijing 100811, China.}
	\thanks{G. Wang and S. Cui are with the State Key Laboratory of Autonomous Intelligent Unmanned Systems, Beijing Institute of Technology, Beijing 100081, China.}
%	\thanks{Ahmed El Shafie is with Apple, California, USA.}
}

\maketitle

\begin{abstract}
	By using an parametric value function to replace the Monte-Carlo rollouts for value estimation, the actor-critic (AC) algorithms can reduce the variance of stochastic policy gradient so that to improve the convergence rate. 
	While existing works mainly focus on analyzing convergence rate of AC algorithms under Markovian noise, the impacts of momentum on AC algorithms remain largely unexplored. 
	In this work, we first propose a heavy-ball momentum based advantage actor-critic (\mbox{HB-A2C}) algorithm by integrating the heavy-ball momentum into the critic recursion that is parameterized by a linear function. 
	When the sample trajectory follows a Markov decision process, we quantitatively certify the acceleration capability of the proposed HB-A2C algorithm.
	Our theoretical results demonstrate that the proposed HB-A2C finds an $\epsilon$-approximate stationary point with $\oo{\epsilon^{-2}}$ iterations for reinforcement learning tasks with Markovian noise. 
	Moreover, we also reveal the dependence of learning rates on the length of the sample trajectory.  
	By carefully selecting the momentum factor of the critic recursion, the proposed HB-A2C can balance the errors introduced by the initialization and the stoschastic approximation. 
\end{abstract}

\begin{IEEEkeywords}
Acceleration, actor-critic algorithms, heavy-ball momentum.
\end{IEEEkeywords}

\section{Introduction}
In model-free reinforcement learning (MFRL) algorithms, an agent optimizes a long-term cumulative reward (a.k.a., value function) by interacting with an unknown stochastic environment that can be articulated as a Markov decision process (MDP).
When combined with the function approximators (e.g., linear approximators and neural networks), the MFRL algorithms have achieved human-level control and extraordinary empirical success in many domains, e.g., 
video games \cite{Mnih2015}, 
robotic control \cite{Ju2022, Wu2022},
autonomous vehicles \cite{Ren2024, Zhang2021}, 
and linear quadratic control tasks \cite{Li2022a, Li2022}.

The current MFRL algorithms can be classified into three categories, i.e., 
\mbox{policy-based} MFRL \cite{Sutton1999, Agarwal2021, Wang2020, Huang2020, Yang2022, Lan2023, schulman2017proximal}, 
\mbox{value-based} MFRL \cite{Bhandari2018, Zou2019, Sun2022, Li2022, Xu2020}, and \mbox{actor-critic} MFRL algorithms \cite{Konda1999, Hong2023}. 
The policy-based MFRL algorithms aim at optimizing the behavior policy based on the policy gradient theorem \cite{Sutton2018}.
When using a parametric policy, the policy-based MFRL can directly optimize the policy parameters via the stochastic gradient descent (SGD) \cite{Sutton1999, Agarwal2021, Wang2020, Huang2020}.
However, the policy-based MFRL algorithms require access to the gradient of the value function with respect to a given policy. 
In practical scenarios with the unknown transition kernels for the MDP, the policy gradients should be estimated from the Monte-Carlo rollouts. 
Consequently, the policy-based MFRL algorithms often encounter significant variance in policy gradients and high sampling costs due to the stochastic approximation.
Besides, the policy-based MFRL algorithms demand for sufficiently small learning rates to guarantee the stable convergence under the function approximators. 
Therefore, the policy-based MFRL algorithms can suffer from a slow convergence.
While appropricte geometry engineering can improve the convergence \cite{schulman2017proximal, Yang2022, Lan2023}, it is still in high demand for reducing the gradient variance of the policy-based MFRL algorithms.

The value-based MFRL algorithms recursively update the the long-term cumulative rewards for different state-action pairs based on the Bellman equation and determine the policy based on the action-value function,~e.g.,~\emph{Q}-learning \cite{Xu2020}. 
Moreover, SARSA can speed the learning process of \emph{Q}-learning by by using the policy improvement operators \cite{Zou2019}.
By estimating the value of successor states via the bootstrapping operation, the value-based MFRL algorithms can efficiently converge to a satisfying behavior policy based on the fixed-point recursions. 
Besides, the value-based MFRL can also be used to evaluate a behavior policy so that to track the future rewards of all states, e.g., temporal-difference (TD) learning algorithms \cite{Bhandari2018}. 
When parameterizing the value function via myopical function approximators, the value-based MFRL algorithms become unstable or diverge for the environments with continuous state and/or action spaces. 
Therefore, an extensive hyperparameter tunning can be required to obtain stable behavior policy when using value-based MFRL algorithms. 
To handle the sample inefficiency and the divergence of the aforementioned MFRL algorithms, recent researches aim at reducing the variance of policy gradient by integrating the policy evalution into the policy improvement so that to propose the actor-critic (AC) algorithms  \cite{Konda1999,  Hong2023, Dai2018, Sutton2018}.
More specficially, the AC algorithms are designed to use a critic recursion to estimate the value of a current policy and then apply an actor recursion to improve the behavior policy based on feedback from the critic \cite{Qiu2021}.

The current AC algorithms can be categorized into \mbox{double-loop} AC algorithms and \mbox{single-loop} AC algorithms. 
In the context of \mbox{double-loop} setting, the critic is consecutively updated for several rounds to obtain an accurate value estimation before each actor recursion \cite{Qiu2021, Kumar2023, Zhang2021}. 
When the actor and critic recursions use different sample trajectories, the inner-loop policy evaluation can be decoupled from the outer-loop policy improvement  \cite{Qiu2021, Kumar2023, Zhang2021}.
Moreover, several different schemes for updating critic sequence have been investigated in centralized topology \cite{Qiu2021, Kumar2023} and decentralized topology \cite{Zhang2021}. 
While mainly utilized for the analytical convenience, the double-loop setting is seldom employed in practice due to the double-sampling requirement for the actor and critic recursions. 
Besides, it is unclear whether an accurate policy evaluation is necessary since it pertains to just one-step policy improvement. 

%the finite-time analysis of actor and critic sequences can be decoupled as the inner-loop policy evaluation and the outer-loop policy improvement subproblems \cite{Qiu2021, Kumar2023, Xu2020a, Zhang2021}. 
%The finite-time convergence of such \mbox{double-loop} AC algorithms has been established under different actor recursion update schemes in \cite{Qiu2021, Kumar2023, Xu2020a} and decentralized setup \cite{Zhang2021}. 

For the \mbox{single-loop} AC algorithms, the actor and critic sequences are updated concurrently \cite{Zhang2020a, Khodadadian2023}. 
The asymptotic convergence of the single-loop AC algorithms has been established from the perspective of ordinary differential equations, specifically when the ratio of the learning rates between the actor and critic approaches zero \cite{Zhang2020a, Khodadadian2023}.
While the asymptotic convergence of single-loop AC algorithms has been well-investigated \cite{Zhang2020a}, the finite-time convergence analysis was unclear until recently \cite{Wang2020, Chen2021, Wu2020, Shen2022, Olshevsky2023, Khodadadian2023}.
In the Big Data era, it is more preferred to use finite-time (or -sample) error bounds towards characterizing the data efficiency of machine learning algorithms. 
For example, by confining the actor sequence to converge slower to the critic sequence, 
the finite-time analysis in \cite{Wu2020} shows that the two-timescale AC algorithm holds a convergence rate $\oo{\nicefrac{1}{K^{0.4}}}$.
The convergence rate of the AC algorithms is sharpened to $\oo{\nicefrac{1}{\sqrt K}}$ when the variance of Markovian noise decays at the same rate as the convergence of critic sequence \cite{Shen2022}. 
The smoothness of the Hessian matrix for the parametric policy is also required to establish the finite-time convergence \cite{Shen2022, Olshevsky2023}.
Moreover, the proposed finite-time convergence analysis in \cite{Olshevsky2023} is only suitable to discrete state-action space and require non-trivial research effort to be extended to the contiuous state-saction space. 
Using the same order of learning rates for the actor and critic sequences, the convergence rate of the single-loop AC algorithm is improved to $\oo{\nicefrac{\log^2 K}{\sqrt K}}$ in \cite{ChenDec.2023}.

\textbf{Contributions.}~Different from \cite{Wu2020, Shen2022, Chen2021, Olshevsky2023, ChenDec.2023}, we consider to improve the convergence of AC algorithms by using momentum. 
More specifically, we introduce the heavy-ball (HB) momentum to the critic recursion and propose the heavy-ball based advantage actor-critic \mbox{(HB-A2C)} algorithm. 
Besides, the actor and critic recursions rely on an Markovian trajectory that are collected from a single MDP in an online manner. 
Our major contributions are summarized as follows. 
\begin{itemize}
	\item For the MFRL tasks, we propose an HB-A2C algorithm that uses a \mbox{$T$-step} trajectory to update the actor and critic parameters. 
	\item We present a new analytical framework that can tightly characterize the estimation error introdued by the gradient bias and the optimality drift under Markovian noise when the heavy-ball based critic recursion is used. 
	Compared with \cite{ChenDec.2023}, our analytical framework can be adopted to characterize the impacts of HB momentum on the convergence. 
	Moreover, our analytical framework demonstrates that the proposed HB-A2C algorithm converges at a rate of $\oo{\nicefrac{1}{\sqrt K}}$ without assuming the decaying variance of Markovian noise. 
\end{itemize}

%Note that the theoretical proofs of the supporting lemmas and theorems are relegated to the online document \cite{} due to the space limitation. 

\textbf{Notation:}
The filtration is denoted by ${\cal F}_k$ that contains all random variables before the start of frame $k$. 
The vector $w^{\dag}$ denotes the transpose of $w$. 
For notational brevity, the parametric distribution  $\pp{ s_{k,t} \in \cdot| s_{k,0}; v_k }$ is denoted by ${\cal P}_{k, t}$.

%Therefore, we can define $\mathds{E}_t[ h(v_k, w_k^*; o_{k,t}) | {\cal F}_k] = \int_{o_{k,t}} {\cal P}_{k,t}\otimes\pi_k\otimes\mathds{P}(o_{k,t}) h(v_k, w_k^*; o_{k,t}) do_{k,t}$. 
%The policy gradient can be recast as $\nabla J(v_k; s_{k,0}) = (1-\gamma)\sum_{t=0}^{\infty}\gamma^t\mathds{E}_t[ h(v_k, w_k^*; o_{k,t}) | {\cal F}_k]$.

\section{Preliminaries}
\subsection{Problem description}
We consider an MDP that is described by a quintuple $({\cal S}, {\cal A}, \mathds{P}, r, \gamma)$, where 
${\cal A}$ is the continuous action space, 
$\cal S$ is the continuous state space, 
$\mathds{P}$ is the unknown transition kernel that maps each state-action pair $(s,a) \in {\cal S}\times {\cal A}$ to a distribution $\pp{\cdot | s, a}$ over state space $\cal S$, 
$r: {\cal S}\times {\cal A} \rightarrow [-R_r, R_r]$ specifies the bounded reward for state-action pair $(s,a)$, and $\gamma$ is the discount factor.

A policy $\pi$ maps state $s$ to a distribution $\pi(\cdot | s)$ over the action sapce $\cal A$.
To evaluate the expected discounted reward starting from a state $s_0$ under the policy $\pi$, the value function is defined as
\begin{equation}\label{eqaa:01a}
	V(s) = \EE{\sum_{k = 0}^{\infty} \gamma^{k} r(s_{k}, a_{k}) \Big| s_0 = s }
\end{equation}
where each action $a_{k}$ follows the policy $\pi(\cdot| s_{k})$, and 
the successor state $s_{k+1} \sim \pp{\cdot | s_{k}, a_{k}}$. 

Given a policy $\pi$, the value function \eqref{eqaa:01a} satisfies the Bellman equation as \cite{Bhandari2018, Sutton2018, Duan2024}
\begin{equation}\label{eqaa:01b}
V(s) = \ee{  r(s,a)  + \gamma^{T} V(s') } 
\end{equation}
where the expectation is taken over the action $a \sim \pi(\cdot| s)$ and the successor state $s' \sim \pp{\cdot | s, a}$.

The objective is to estimate the optimal policy $\pi^*$ so that to maximize the expected discounted reward $J(\pi)$ as
\begin{equation}\label{eqaa:01c}
\pi^* \in \arg\max J(\pi):= (1-\gamma)\ee{ V(s) }.
\end{equation}

\subsection{Function approximation}
When considering the continuous state and action spaces, it becomes computational burdensome to obtain the optimal policy $\pi^*$ or even intractable due to the notorious issue of curse of dimensionality (CoD). 
One popular way to handle the CoD issue is to approximate each policy $\pi_k$ and the value function $V(s)$ by a neural network and a linear-function approximator, respectively.
In this work, the policy $\pi(a|s)$ and the value function $V(s)$ are respectively parameterized by the actor parameter $v \in \mathbb{R}^{d_v}$ and the critic parameter $w \in \mathbb{R}^{d_w}$.
More specficially, the parametric policy is denoted by $\pi(a|s) = \pi_{v}(a|s)$, and the parametric value function is denoted by $V(s) \approx V_w(s) = \phi^{\dag}(s) w$ with $\norm{w} \le R_w$ and the feature embedding $\phi(s)$ satisfying $\norm{\phi(s)} \le 1$, $s \in {\cal S}$. 
Note that the optimal value $V^*(s) \equiv V_{w^*}(s)$ when the radius $R_w$ is sufficient large \cite{Bhandari2018}.

Based on the parametric policy $\pi_v$, parametric value function $V_w$, and the Bellman equation \eqref{eqaa:01b}, we can recast the objective in \eqref{eqaa:01c} as a bilevel optimization that optimize the actor parameter $v$ in the outer problem and the critic parameter $w$ in the innter problem as 
\begin{equation}\label{eqaa:02}
\begin{split}
v^* = \arg\max_{v}&~ J(v)  \\
 \mbox{s.t.}&~w^* \!=\! \arg\min_{\norm{w}\le R_w}  \frac{1}{2}\ee{\norm{ V_w^{\tar}(s) \!-\! V_w(s) }^2} 
\end{split}
\end{equation}
where $J(v) := \ee{J(v, w^*; s)}$ with $J(v, w^*; s) := (1-\gamma)V_{w^*}(s)$, and $V_w^{\tar}(s)$ is the target value for state $s$. 
The target value $V_w^{\tar}(s)$ can be estimated by the one-step (or multi-step) bootstrapping. 
\begin{remark}
According to the inner problem of \eqref{eqaa:02}, the optimal critic parameter $w^*$ is essentially a function of the actor parameter $v$. 
Therefore, the only optimization variable is the actor parameter $v$ in the outer problem of \eqref{eqaa:02}. 
\end{remark}

\section{Heavy-Ball Based Actor-Critic for RL Tasks}
\subsection{Algorithm development}
We consider a fully data-driven technique that maintains a running estimate of the value function (cf. the inner problem in \eqref{eqaa:02}) while performing policy updates based on the estimated state values (cf. the outer problem in \eqref{eqaa:02}). A multi-step bootstrapping is employed to estimate the target value $V_w^{\text{tar}}(s)$. One of the merits of multi-step bootstrapping is the ability to balance bias and variance during the estimation of the target value. Furthermore, as we will justify later, the multi-step bootstrapping allows for a larger learning rate when solving the inner problem of \eqref{eqaa:02} using recursive updates, thereby reducing the number of recursions required for the critic parameter. Consequently, we consider the MDP to operate on two timescales, where each coarse-grain slot (i.e., frame) consists of $T$ fine-grain slots (i.e., $T$ steps). 
For notational brevity, we recast 
the reward $r(s_{k,t}, a_{k,t})$, 
the feature embedding $\phi(s_{k,t})$, 
the policy $\pi_{v_k}$, and 
the the optimal critic $w^*_{v_k} $ as 
$r_{k,t} := r(s_{k,t}, a_{k,t})$, 
$\phi_{k,t} := \phi(s_{k,t})$, 
$\pi_{v_k} := \pi_k$, and $w^*_{v_k} := w^*_k$, respectively.

\textbf{Inner optimization.}~For the inner optimization, the critic parameter $w$ can be updated via the  stochastic semi-gradient
\begin{equation}\label{eqaa:05a}
	g(w_k; O_{k}) = [V_{w_k}^{\tar}(s_{k,0}) - V_{w_k}(s_{k,0})]\nabla V_{w_k}(s_{k,0}) 
\end{equation}
where the parametric value $V_{w_k}(s_{k,0}) = \phi^{\dag}_{k, 0} w_k$; the target value $V_{w_k}^{\tar}(s_{k,0})$ is estimated by a \mbox{$T$-step} bootstrapping as $V_{w_k}^{\tar}(s_{k,0}) = \sum_{t=0}^{T-1}\gamma^t r_{k,t} + \gamma^T \phi^{\dag}_{k,T} w_k$; $O_{k} = [o_{k,t}]_{t=0}^{T-1}$ is the $T$-step trajectory; and the observation  $o_{k,t} = (s_{k,t}, a_{k,t}, s_{k,t+1})$ follows the distribution ${\cal P}_{k,t} \otimes \pi_k \otimes \mathds{P}$ with  ${\cal P}_{k,t} := \pp{s_{k,t} \in \cdot | s_{k,0}; v_k}$ as the $t$-step transition kernel.

The compact form stochastic semi-gradient is denoted by 
\begin{equation}\label{eqaa:05}
g(w_k; O_{k}) = \Phi_{k} w_k - b_{k}
\end{equation}
where $\Phi_{k} =\! \phi_{k, 0}[\phi_{k, 0} - \gamma^T \phi_{k, T}]^{\dag}$ and $b_{k} =\! \phi_{k, 0} \sum_{t=0}^{T-1}\gamma^t r_{k, t}$. 

The stochastic semi-gradient \eqref{eqaa:05} equals to the sum of full semi-gradient and gradient bias as
\begin{equation}\label{eqaa:06}
	g(w_k; O_{k}) = \ee{g(w_k; \bar O_k)} + \zeta(v_k, w_k; O_k)
\end{equation}
where the gradient bias is $\zeta(v_k, w_k; O_k) := \Phi_{k} w_k - b_{k} - \ee{g(w_k;  \bar O_k)}$, and the full semi-gradient $\ee{g(w_k;  \bar O_k)} = \ee{\bar\Phi_k w_k - \bar b_k }$ with $\bar\Phi_{k} = \bar\phi_{k, 0}[\bar\phi_{k, 0} - \gamma^T \bar\phi_{k, T}]^{\dag}$ and $\bar b_k =  \bar\phi_{k,0}\sum_{t=0}^{T-1}\gamma^t \bar r_{k,t}$.
Denoting the $\pi_k$-induced stationary distribution by $\mu_k$, the $T$-step sample trajectory $\bar O_k$ is obtained from the stationary distribution $\mu_k \otimes \pi_k \otimes \mathds{P}$.

Note that, given the actor parameter $v_k$, the optimal critic paramter $w_k^*$ satisfies $\ee{g(w_k^*; \bar O_k)} = 0$. 
Together with the full semi-gradient $\ee{g(w_k;  \bar O_k)} = \ee{\bar\Phi_k w_k - \bar b_k }$, we obtain 
\begin{equation}\label{eqaa:13}
	\ee{g(w_k; \bar O_k)} - \ee{g(w_k^*; \bar O_k)} 
	= \ee{\bar\Phi_{k}(w_k - w_k^*)}.
\end{equation}

Recalling the definitions of $\bar\Phi_k$ and $V_{w_k}(s_{k,0})$ and setting  $\varepsilon_1 = V_{w_k}(\bar s_{k,0}) - V_{w_k^*}(\bar s_{k,0})$ and $\varepsilon_2 = V_{w_k}(\bar s_{k,T}) - V_{w_k^*}(\bar s_{k,T})$, we have $\ee{\bar\Phi_{k}(w_k - w_k^*)} = \ee{\bar\phi_{k,0}(\varepsilon_1 - \gamma^T\varepsilon_2)}$. 
Based on \cite[Lemma 3]{Bhandari2018}, we obtain
\begin{equation}\label{eqaa:14}
	\begin{split}
		{\inp{w_k - w_k^*, \ee{\bar\Phi_{k}(w_k - w_k^*)}}} 
		&= \ee{\varepsilon_1^2}  - \gamma^T\ee{\varepsilon_1\varepsilon_2} \\
		&\ge \sigma \norm{w_k - w_k^*}^2
	\end{split}
\end{equation}
where $\sigma := (1-\gamma^T)\lambda$ with $\lambda > 0$ as the smallest eigenvalue of the  matrix $\int_{\mu_k(s)}\phi(s)\phi^{\dag}(s) \mu_k(s) ds$. When the redundant or the irrelevant features are removed, the matrix $\int_{\mu_k(s)}\phi(s)\phi^{\dag}(s) \mu_k(s) ds$ is \mbox{positive-definite}. 
Since each feature embedding $\phi(s)$ satisfies $\norm{\phi(s)} \le 1$, the smallest eigenvalue satisfies $\lambda \in (0, 1)$ \cite{Bhandari2018}.

\begin{remark}
The inequality in \eqref{eqaa:14} can be viewed as a strongly monotone property of the full semi-gradient $\ee{g(w_k; \bar O_k)}$.
Based on \eqref{eqaa:14}, we observe that a longer trajectory $\bar{O}_k$ results in a larger condition number $\sigma$ for the inner problem in \eqref{eqaa:02}, thereby allowing for a higher learning rate so that to reduce the number of training recursions required for the critic parameter.
\end{remark}

\textbf{Outer optimization.}~The policy gradient theorem \cite{Sutton1999} provides an analytical experession for the gradient of outer objective in \eqref{eqaa:02}. 
In the context of two timescale framework, the policy gradient of $J(v_k)$ is defined as $\nabla J(v_k) = \ee{ \nabla J(v_k, w_k^*; s_{k,0}) }$. 
Moreover, the policy gradient of frame $k$ is given by 
\begin{equation}\label{eqaa:03}
\nabla J(v_k, w_k^*; s_{k,0}) = (1-\gamma)\sum_{t=0}^{\infty}\gamma^t \ee{ h(v_k, w_k^*; o_{k,t}) }
\end{equation}
where the expectation is taken over each observation $o_{k,t} \sim {\cal P}_{k,t} \otimes \pi_k \otimes \mathds{P}$, and the function $h(v_k, w_k^*; o_{k,t})$ is defined as 
\begin{equation}\label{eqaa:04}
\begin{split}
&h(v_k, w_k^*; o_{k,t}) \\
&\!:=\! [r_{k,t} + [\gamma\phi_{k, t+1} - \phi_{k,t}]^{\dag}w_k^*]\nabla\log\pi_k (a_{k,t} | s_{k,t})
\end{split}
\end{equation}
where $w_k^*$ is the optimal value parameter under the policy $\pi_k$.

Based on \eqref{eqaa:05} and \eqref{eqaa:03}, the vanilla SGD can be used to search for the optimal actor parameter $v^*$ and the optimal critic parameter $w^*$. 
However, the vanilla SGD can suffer from slow convergence. 
Therefore, we leverage the HB momentum to improve the convergence rate and propose an HB based advantage actor-critic (HB-A2C) algorithm in \textbf{Algorithm \ref{alg:01}}.

\begin{algorithm}[htb]\small
	\centering
	\caption{HB-A2C Algorithm}\label{alg:01}
	\begin{algorithmic}[1]
		\State \textbf{Initialization:} 
		critic hyper-parameters: stepsize $\beta$, parameter $w_0$, momentum factor $\eta_1$, and momentum parameter $n_{-1} = 0$; 
		actor hyper-parameters: stepsize $\alpha$ and parameter $v_0$
		\For{$k = 0, 1, \ldots, K-1$}
		\State Rolling out $T$-step observations $O_{k} = [o_{k,t}]_{t=0}^{T-1}$ via the behavior $\pi_{k}$ \label{sampling}
		\State Update the critic parameter based on \eqref{eqaa:05} as 
		\begin{subequations}\label{eqaa:07}
			\begin{align}
				n_k &=    (1-\eta_1) n_{k-1} + \eta_1 g(w_k; O_{k}) \label{eqaa:07a}\\
				w_{k+1} &=  \proj_{R_w}[ w_{k} - \beta n_k  ] \label{eqaa:07b}
			\end{align}
		\end{subequations}
		where the operator $\prod_{R_w}[\cdot]$ denotes the projection onto the region $\norm{w} \le R_w$ \label{critic:end}
		\State Calculate the stochastic policy gradient based on \eqref{eqaa:04} as 
		\begin{equation}\label{eqaa:08}
		H(v_k, w_k; O_k) = (1-\gamma)\sum_{t=0}^{T-1} \gamma^t h(v_k, w_k; o_{k,t})
		\end{equation}
		\State Update the actor parameter based on \eqref{eqaa:08} as 
%		\begin{subequations}\label{eqaa:09}
%		\begin{align}
%		u_k &= (1 - \eta_2) u_{k-1} + \eta_2 H(v_k, w_k; O_k) \label{eqaa:09a}\\
%		v_{k+1} &= v_k + \alpha u_k  \label{eqaa:09b}
%		\end{align}
%		\end{subequations}
		
		\begin{equation}\label{eqaa:09}
		v_{k+1} = v_k + \alpha H(v_k, w_k; O_k)
		\end{equation}
		\EndFor
	\end{algorithmic}
\end{algorithm}

\subsection{Convergence analysis}
Hereinafter, our goal is to analyze the convergence rate of the proposed HB-A2C algorithm for a realistic setting where the transitions are sampled along a trajectory of the MDP.
To proceed, we need the ensuing assumptions on behavior to facilitate our analysis.

\begin{assump}\label{assum:01}
For each state-action pair $(s, a) \in ({\cal S, A})$, the behavior policy satisfies
\begin{subequations}\label{eqaa:10}
\begin{align}
&\norm{ \nabla\log\pi_v(a|s) } \le R_\pi \label{eqaa:10a}\\
&\norm{ \nabla\log\pi_v(a|s) \!-\! \nabla\log\pi_{v'}(a|s) } \!\le\! L_\pi'\norm{v - v'} \label{eqaa:10b}\\
&|\pi_v(a|s) - \pi(a|s)| \le L_\pi \norm{v - v'} \label{eqaa:10c}
\end{align}
\end{subequations}
where $v, v' \in \mathbb{R}^{d_v}$.
\end{assump}

Assumption \ref{assum:01} is standard for the analysis of policy gradient based methods, see e.g., \cite{Zou2019, Hong2023, Wang2020, ChenDec.2023, Wu2020, Shen2022, Olshevsky2023, Kumar2023}. 
The Lipschitz continuity assumption holds for canonical parametric policies, such as, the Gaussian policy \cite{Doya2000} and Boltzman policy \cite{Konda1999a}. 
Assumption \ref{assum:01} guarantees that the expected discounted reward $J(v)$ has $L$-Lipschitz continuous gradient as
\begin{equation}\label{eqaa:11}
\inp{ \nabla J(v), v'-v }  
\le J(v') - J(v) + \frac{1}{2}L\norm{v'-v}^2
\end{equation}
where $v \in \mathbb{R}^{d_v}$ and  $v' \in \mathbb{R}^{d_v}$. The detailed derivations are relegated to Lemma \ref{lemmas:06} in Appendix.

\begin{assump}\label{assum:02}
For each behavior $\pi_k$, the induced Markov chain is ergodic and has a stationary distribution $\mu_k$ with $\mu_k(s) > 0$, $s \in \cal S$. Moreover, there exist contants $c_0$ and $\rho \in (0, 1)$ such that 
\begin{equation}\label{eqaa:12}
\normtv{{\cal P}_{k,t} - \mu_k} \le c_0 \rho^t
\end{equation}
where the total variation distance for the two probability meatures ${\cal P}_{k,t}$ and $\mu_k$ is defined as $\normtv{{\cal P}_{k,t} - \mu_k} = \int_{s} |{\cal P}_{k,t}(s) - \mu_k(s) | ds$.
\end{assump}

The first part of Assumption \ref{assum:02} (i.e., ergodicity) ensures that all states are visited an infinite number of times and  the existence of a mixing time for the MDP.
The second part (i.e., the mixing time of the policy in \eqref{eqaa:12}) guarantees that the optimal policy can be obtained from a single sample trajectory of the MDP.
It is worth remarking that Assumption \ref{assum:02} is a standard requirement for theoretical analysis of the RL algorithms; see e.g., \cite{Khodadadian2023, ChenDec.2023, Wu2020, Bhandari2018, Levin2017}.

Before characterizing the convergence properties of the proposed HB-A2C algorithm, we start by establishing several lemmas.

\begin{lemma}\label{lemmas:01}
	The (stochastic) semi-gradient of critic and (stochastic) policy gradient of actor are bounded as
	\begin{subequations}\label{eqaa:15}
		\begin{align}
			& \norm{g(w_k; O_k)} \le R_g, \norm{\ee{g(w_k;\bar O_k)}} \le R_g \label{eqaa:15a}\\
			& \norm{H(v_k, w_k; O_k)} \le R_h, \norm{\nabla J(v_k)} \le R_h \label{eqaa:15b}
		\end{align}
	\end{subequations}
	where $R_g = (1+\gamma^T) R_w  + c_1(\gamma)R_r$ and $R_h = R_\pi[R_r \!+\! (1+\gamma)R_w]$ with 
	$c_1(\gamma) = \nicefrac{(1 - \gamma^T)}{(1-\gamma)}$.
\end{lemma}
\begin{IEEEproof}
See Appendix \ref{les:01}.
\end{IEEEproof}

Lemma~\ref{lemmas:01} provides bounds for both the (stochastic) semi-gradient of the critic and the (stochastic) policy gradient of the actor that are useful for controlling (i.e., upper-bounding) the drifts of the critic and actor parameters as follows.

The recursion in \eqref{eqaa:07a} can be recast as $n_k = \eta_1 \sum_{\tau=0}^k (1-\eta_1)^{k - \tau} g(w_\tau; O_\tau)$ when $n_{-1} = 0$. Therefore, the upper bound of $\norm{n_k}$ is derived as 
\begin{equation}\label{eqaa:16}
\norm{n_k} = \NORM{ \eta_1 \sum_{\tau=0}^k (1-\eta_1)^{k - \tau} g(w_\tau; O_\tau) } \le R_g.
\end{equation}

Based on the recursion in \eqref{eqaa:07b} and \eqref{eqaa:16}, we obtain the one-frame drift of critic as 
\begin{equation}\label{eqaa:17}
	\norm{w_{k+1} - w_k}  \le \norm{\beta n_k } \le  R_g\beta
\end{equation}
where the first inequality follows from the non-expansive property of the projection operation.

%The recursion in \eqref{eqaa:09a} can be recast as $u_k = \eta_2\sum_{\tau = 0}^k (1-\eta_2)^{k-\tau} H(v_\tau, w_\tau; O_\tau)$ when $u_{-1} = 0$.
%Hence, the upper bound of $\norm{u_k}$ is 
%\begin{equation}\label{eqaa:18}
%\norm{u_k} = \NORM{ \eta_2\sum_{\tau = 0}^k (1-\eta_2)^{k-\tau} H(v_\tau, w_\tau; O_\tau) } \le R_h.
%\end{equation}

Based on \eqref{eqaa:09} and Lemma \ref{lemmas:01}, we obtain the one-frame drift of actor as 
\begin{equation}\label{eqaa:19}
\norm{v_{k+1} - v_k} \le R_h \alpha. 
\end{equation}

Based on Lemma \ref{lemmas:01} and \eqref{eqaa:17}, we can investigate the properties of the gradient bias term $\zeta(v_k, w_k; O_{k})$. 

\begin{lemma}\label{lemmas:02}
	Suppose Assumptions \ref{assum:01} and \ref{assum:02} hold. 
	When length of each trajectory satisfies $T \ge \nicefrac{\log c_0^{-1} \beta}{\log\rho}$, 
	the gradient bias $\zeta(v_k, w_k; O_k)$ satisfies
	\begin{equation}\label{eqaa:20}
		\norm{\zeta(v_k, w_k; O_{k}) - \zeta(v_k, w_{k-1}; O_{k}) } \le 8 R_g \beta
	\end{equation}
	and 
	\begin{equation}\label{eqaa:21}
		\begin{split}
&\norm{\ee{ \zeta(v_k, w_{k-1}; O_k) | {\cal F}_{k-1}} } \\
&\le  (c_2 + 2T)|{\cal A}|L_\pi R_g\norm{v_k - v_{k-1}} + R_g\beta
		\end{split}
	\end{equation}
	where $c_2 > 0$, and ${\cal F}_{k-1}$ denotes the filtration that contains all randomness prior to frame $k-1$.
\end{lemma}
\begin{IEEEproof}
See Appendix \ref{les:02}.
\end{IEEEproof}

Lemma \ref{lemmas:02} implies that: 1) the one-frame drift of gradient bias $\zeta(v_k, w_k; O_k)$ with respect to $w_k$ can be confined by the critic stepsize $\beta$; and 2) the gradient bias does not rapidly increase with $\norm{v_k - v_{k-1}}$, which serves as one of the keys in developing our subsequent convergence of the critic sequence of the HB-A2C algorithm.

Based on \eqref{eqaa:05}, we observe that the optimal critic $w_k^*$ per frame $k$ is a function of $v_k$.
Therefore, we are motivated to investigate the drift of optimal critic with respect to the actor parameters.

\begin{lemma}\label{lemmas:05}
	When Assumption \ref{assum:01} is satisfied, the optimal critic parameter $w^*$ per frame satisfies
	\begin{align}
		\norm{w^*_v - w^*_{v'}} &\le L_*\norm{v-v'} \\
		\norm{ \nabla w^*_v } &\le G_*
	\end{align}
	where $L_* > 0$, $G_* > 0$, $v \in \mathbb{R}^{d_v}$, and  $v' \in \mathbb{R}^{d_v}$.
\end{lemma}
\begin{IEEEproof}
	See Appendix \ref{les:05}.
\end{IEEEproof}

Lemma \ref{lemmas:05} shows that the drift of the optimal critic is controlled by the drift of the actor. 
Before analyzing the convergence behavior of the actor recursion \eqref{eqaa:09}, we need to establish the Lipschitz continuity of the stochastic policy gradient $H(v, w; O_k)$ with respect to the critic parameter $w$. 

\begin{lemma}\label{lemmas:03}
	When Assumption \ref{assum:01} is satisfied, the stochastic policy gradient in \eqref{eqaa:08} is Lipschitz with respect to $w$ as 
	\begin{equation}\label{eqaa:22}
		\norm{H(v, w; O_k) - H(v, w'; O_k)} 
		\le (1 + \gamma)R_\pi \norm{w - w'}
	\end{equation}
	where $w \in \mathbb{R}^{d_w}$ and $w' \in \mathbb{R}^{d_w}$
\end{lemma}
\begin{IEEEproof}
	See Appendix \ref{les:03}.
\end{IEEEproof}

Based on Lemma \ref{lemmas:03}, we now present the convergence behavior of the policy gradient $\nabla J(v_k)$ as follows. 

\begin{theorem}\label{thm:02}
	Suppose Assumptions  \ref{assum:01} and \ref{assum:02} hold, and set $u_{-1} = u_0 = 0$.
	When the minibatch size $T \ge \nicefrac{\log \beta}{2\log\gamma}$, the $K$-step convergence of actor is
	\begin{equation}\label{eqaa:24}
		\begin{split}
			&\frac{\alpha}{2 K}\sum_{k=0}^{K-1}\!\norm{\nabla J(v_k)}^2 \!-\! \frac{(1+\gamma)^2 R_\pi^2 \alpha}{K}\sum_{k=0}^{K-1}\!\ee{ \norm{\Delta_k}^2 } \\
			&\le \frac{1}{K} [J(v_K) - J(v_0)] 
			+ \frac{1}{2} L R_h^2  \alpha^2 + R_h^2 \alpha\beta
		\end{split}
	\end{equation}
	where $\Delta_k = w_k - w_k^*$.
\end{theorem}
\begin{IEEEproof}
	The finite-time convergence analysis of actor starts from that the expected discounted reward $J(v)$ under state $s$ has the $L$-Lipschitz continuous gradient. 
	Together with the recursion in \eqref{eqaa:09} and the inequality \eqref{eqaa:19}, we have
	\begin{equation}\label{eqaa02:01}
		\alpha \ee{\Omega_k } \le {J}(v_{k+1}) - {J}(v_{k}) + \frac{1}{2} L R_h^2  \alpha^2
	\end{equation}
	where $\Omega_k = \inp{ \nabla J(v_k), H(v_k, w_k; O_k) }$.
	
	Based on the definitions of policy gradient $\nabla J(v_k) = \ee{ \nabla J(v_k, w_k^*;s_{k,0}) }$ with $\nabla J(v_k, w_k^*;s_{k,0}) = (1-\gamma)\sum_{t=0}^{\infty} \gamma^t \int_{o_{k,t}} {\cal P}_{k,t}\otimes\pi_k\otimes\mathds{P}(o_{k,t}) h(v_k, w_k^*; o_{k,t}) d o_{k,t}$ and the stochastic policy gradient $H(v_k,w_k^*; O_k) =  (1-\gamma)\sum_{t=0}^{T-1} \gamma^t h(v_k, w_k^*; o_{k,t})$, we have 
	\begin{equation}\label{eqaa02:04}
		\begin{split}
			&\inp{ \nabla J(v_k), \ee{H(v_k, w_k; O_k)} } \\
			&\ge \frac{1}{2}\norm{ \nabla J(v_k) }^2 - \frac{1}{2}\norm{ \nabla J(v_k) - \ee{H(v_k, w_k; O_k)} }^2\\
			&\ge \frac{1}{2}\norm{\nabla J(v_k)}^2 
			-  R_h^2 \beta 
			- (1 + \gamma)^2R_\pi^2 \ee{ \norm{\Delta_k}^2 }.
		\end{split}
	\end{equation}
	
	Substituting \eqref{eqaa02:04} into \eqref{eqaa02:01}, we obtain
	\begin{equation}\label{eqaa02:05}
		\begin{split}
			&\frac{1}{2}\norm{\nabla J(v_k)}^2  	- (1+\gamma)^2 R_\pi^2\ee{ \norm{\Delta_k}^2 } \\
			& \le {J}(v_{k+1}) - {J}(v_{k}) 
			+ R_h^2\beta 
			+ \frac{1}{2} L R_h^2 \alpha^2.
		\end{split}
	\end{equation}
	
	Summing \eqref{eqaa02:05} over $k = 0, 1, \ldots, K-1$, we complete the proof by obtaining \eqref{eqaa:24}. 
	More detailed information can be found in Appendix \ref{sec:vi}. 
\end{IEEEproof}

We observe from Theorem \ref{thm:02} that the convergence behaviors of policy gradient and critic parameter are coupled.
Therefore, we need to investigate the convergence behavior of the critic parameter so that to establish a unified convergence of both actor and critic recursions. 
Based on Lemmas \ref{lemmas:01}--\ref{lemmas:03}, we can formally present the convergence of the critic update in \eqref{eqaa:07} as follows.

\begin{theorem}\label{thm:01}
	Suppose Assumptions  \ref{assum:01} and \ref{assum:02} hold, and set $v_0 = 0$.
	When the minibatch size $T \ge  \max\{\nicefrac{\log c_0^{-1} \beta}{\log\rho}, \nicefrac{\log \beta}{2\log\gamma}\}$, the $K$-step convergence of critic is
	\begin{equation}\label{eqaa:23}
\begin{split}
	&\frac{1}{K}\Big[\sigma \beta - [(1+\gamma)R_\pi G_* + 2G_*^2]\alpha\Big]\sum_{k=0}^{K-1}\ee{ \norm{\Delta_k}^2 }  \\
	&\le \frac{\alpha}{4K}\sum_{k=0}^{K-1}\norm{\nabla J(v_k)}^2
	\!+\! \frac{1}{2 K}[\ee{\norm{\Delta_0}^2} \!-\! \ee{\norm{\Delta_K}^2}] \\
	&\hspace{0.3cm} + L_*^2 R_h^2 \alpha^2 
	+ [\frac{c_4}{\eta_1} + R_g^2 ]\beta^2  \\
	&\hspace{0.3cm} + [\frac{c_3}{\eta_1} + 2 R_w G_*] R_h \alpha\beta  
	+ \frac{2(1-\eta_1) \beta}{\eta_1 K} R_w R_g
\end{split}
	\end{equation}
	where $c_3 = [(1+\eta_1)L_* + 2\eta_1(c_2 + 2T)|{\cal A}|L_\pi R_w]R_g$ and $c_4 = [2\eta_1(R_g + 9R_w) + (1-\eta_1)R_g]R_g$.
\end{theorem}
\begin{IEEEproof}
The major challenge of analyzing the finite-time convergence of critic comes from chacterizing errors that are related to the gradient variance, optimality drift, and the gradient progress terms as 
\begin{equation}\label{eqaa01:01}
	\begin{split}
		&\frac{1}{2}\norm{\Delta_{k+1}} - \frac{1}{2}\norm{\Delta_k}^2  \\
		&\le \frac{1}{2}\norm{ \beta n_k +  w_{k+1}^* - w_{k}^* }^2 \hspace{0.2 cm}\mbox{(gradient variance)}  \\
		&\hspace{0.3 cm} + \inp{ \Delta_k, w_k^* - w_{k + 1}^* } \hspace{0.84 cm}\mbox{(optimality drift)}\\
		&\hspace{0.3 cm} - \beta \inp{\Delta_k, n_k}.  \hspace{1.88 cm}\mbox{(gradient progress)}\\
	\end{split}
\end{equation}

Since the proposed HB-A2C algorithm integrates the HB momentum into the critic update, our used techniques are different from \cite{ChenDec.2023, Huang2020, Wu2020} when bounding the gradient variance, optimality drift, and gradient progress in \eqref{eqaa01:01}.

\textbf{Step 1: Characterization of gradient variance.}
Recalling that $n_k = \eta_1\sum_{\tau=0}^k (1 - \eta_1)^{k - \tau} g(w_\tau; O_\tau)$ when  $n_{-1} = 0$. 
Together with Lemmas \ref{lemmas:01} and \ref{lemmas:04}, we can upper-bound the gradient variance in \eqref{eqaa01:01} as
\begin{equation}\label{eqaa01:02}
	\frac{1}{2}\norm{ \beta n_k +  w_{k+1}^* \!-\! w_{k}^* }^2 \le R_g^2 \beta^2 + L_*^2 \norm{v_{k+1} - v_k}^2.
\end{equation}

\textbf{Step 2: Characterization of optimality drift.}
Based on Lemmas \ref{lemmas:05}--\ref{lemmas:04}, we can upper-bound the optimality drift as 
\begin{multline}\label{eqaa01:03}
\ee{\inp{ \Delta_k,  w^*_k -  w^*_{k+1} } } 
\le [(1+\gamma)R_\pi G_* + 2G_*^2]\alpha\ee{ \norm{\Delta_k}^2 } \\
+ \frac{1}{4}\alpha\norm{\nabla J(v_k)}^2 + 2\alpha \beta  R_w R_h G_*. 
\end{multline}

Substituting \eqref{eqaa01:02} and \eqref{eqaa01:03} into \eqref{eqaa01:01} and recalling the fact in \eqref{eqaa:19}, we obtain
\begin{equation}\label{eqaa01:04}
\begin{split}
	&\beta\ee{ \inp{\Delta_k, n_k} } -  [(1+\gamma)R_\pi G_* + 2G_*^2]\alpha \ee{\norm{\Delta_k}^2} \\
	&\le \frac{1}{2}[\ee{\norm{\Delta_k}^2} - \ee{\norm{\Delta_{k+1}}^2}] 
	+ \frac{1}{4}\alpha\norm{\nabla J(v_k)}^2  \\
	&\hspace{0.3cm}+ 2\alpha\beta R_w R_h G_* 
	\!+\! R_g^2 \beta^2 
	\!+\! L_*^2 R_h^2 \alpha^2. \!
\end{split}
\end{equation}

\textbf{Step 3: Characterization of gradient progress.} 
The most challenging part locates at analyzing the gradient progress that needs to consider the HB momentum update of the critic. 
More specifically, we can decompose the gradient progress term based on \eqref{eqaa:07a} as 
\begin{equation}\label{eqaa01:05}
	\begin{split}
		& \inp{\Delta_{k}, n_k} \\
		&= (1 \!-\! \eta_1)\inp{\Delta_{k-1}, n_{k-1}}  \\
		&\hspace{0.5 cm} \!+\! (1 \!-\! \eta_1)\inp{\Delta_k \!-\! \Delta_{k-1}, n_{k-1}} 
		\!+\! \eta_1\inp{\Delta_k, g(w_k; O_k)}. 
	\end{split}
\end{equation}

Following the Lipschitz continuity of the gradient bias $\zeta(v_k, w_k; O_k)$ in Lemma \ref{lemmas:02} and the optimal critic parameter $w^*(v)$ in Lemma \ref{lemmas:05} as well as the recursion \eqref{eqaa:07} and the decomposed gradient progress \eqref{eqaa01:05}, we respectively obtain the lower and upper bound of $\eta_1\inp{\Delta_k, g(w_k; O_k)}$ as
\begin{subequations}\label{eqaa01:06}
	\begin{align}
		&\eta_1 \ee{ \inp{\Delta_k, g(w_k; O_k)} } \nonumber\\
		&\ge \eta_1 \sigma \ee{ \norm{\Delta_k}^2 } 
		- \eta_1 c_3'\ee{\norm{v_k - v_{k-1}}}
		- \eta_1 c_4'\beta \label{eqaa01:06a}\\
		&\eta_1 \ee{ \inp{\Delta_k, g(w_k; O_k)} } \nonumber\\
		&\le (1-\eta_1) R_g [R_g \beta + L_*\ee{ \norm{v_k - v_{k-1}} }] \label{eqaa01:06b}\\
		&\hspace{0.3 cm}+ \ee{ \inp{\Delta_{k}, n_k}  } - (1-\eta_1)\ee{ \inp{\Delta_{k - 1}, n_{k-1}}  } \nonumber
	\end{align}
\end{subequations}
where $c_3' := 2R_g [ L_* + (c_2 + 2T)|{\cal A}|L_\pi R_w ]$ and $c_4' := 2R_g(R_g + 9 R_w)$.

Substituting \eqref{eqaa01:06a} into \eqref{eqaa01:06b}, we obtain 
\begin{equation}\label{eqaa01:07}
	\begin{split}
		&\eta_1 \sigma \ee{ \norm{\Delta_k}^2 } \\
		&\le c_3 \ee{ \norm{v_k - v_{k-1}} }
		+ c_4 \beta\\
		&\hspace{0.3 cm}+ \ee{ \inp{\Delta_{k}, n_k}  } - (1-\eta_1)\ee{ \inp{\Delta_{k - 1}, n_{k-1}}  }
	\end{split}
\end{equation}
where $c_3 := [(1+\eta_1)L_* + 2\eta_1(c_2 + 2T)|{\cal A}|L_\pi R_w]R_g$ and $c_4 := [2\eta_1(R_g + 9R_w) + (1-\eta_1)R_g]R_g$.

Summing \eqref{eqaa01:07} over $k = 0, \ldots, K-1$ and recalling the fact $n_{-1} = 0$, we obtain
\begin{multline}\label{eqaa01:007}
\sigma\beta \sum_{k=0}^{K-1}\ee{ \norm{\Delta_k}^2 } 
\le \frac{c_3}{\eta_1} R_h \alpha\beta K 
+ \frac{c_4}{\eta_1} \beta^2 K \\
+ \frac{2(1-\eta_1)}{\eta_1} R_w R_g \beta  
+ \beta\sum_{k=0}^{K-1}\ee{ \inp{\Delta_{k}, n_{k}}  }
\end{multline}

Combining \eqref{eqaa01:04} and \eqref{eqaa01:007} and summing over $k = 0, \ldots, K-1$, we complete the proof by obtaining \eqref{eqaa:23}. 
The detailed derivations can be found in Appendix  \ref{sec:v}.
\end{IEEEproof}

%\begin{figure*}[b]
%	\hrulefill
%	\begin{subequations}\label{eqaa:26}
%		\begin{align}
%			&\frac{\alpha}{2 K}\sum_{k=0}^{K-1}\!\norm{\nabla J(v_k)}^2 + \Big[ \frac{ \sigma\beta }{2 K} - \frac{(1+\gamma)^2 R_\pi^2 \alpha}{K} \Big]\sum_{k=0}^{K-1}\!\ee{ \norm{\Delta_k}^2 } \\
%			&\le  \frac{{\cal L}_K - {\cal L}_0}{ K}
%			\!+\! \Big[ [1 + \frac{1}{2\sigma}] L_*^2 + [1 - \eta_2]L \Big]R_h^2 \alpha^2
%			\!+\! \Big[ \frac{c_3}{\eta_1} + R_h \Big] R_h \alpha \beta
%			\!+\! [\frac{c_4}{\eta_1} + R_g^2] \beta^2
%			\!+\! \frac{2\beta(1-\eta_1)}{\eta_1 K} R_w R_g 
%			\!+\! \frac{2 - \eta_2}{2K} R_h^2 \alpha
%		\end{align}
%	\end{subequations}
%\end{figure*}

Based on Theorems \ref{thm:02} and \ref{thm:01}, we observe that the convergence of the policy gradient $\nabla J(v_k)$ and the critic parameter $w_k$ are coupled with each other. 
By combining Theorems \ref{thm:02} and \ref{thm:01}, we can now establish the unified convergence of the actor and critic recursions as follows.

\begin{corollary}\label{cor:01}
Suppose Assumptions \ref{assum:01} and \ref{assum:02} hold. 
Set $\alpha = \Theta(\nicefrac{1}{\sqrt K})$ and $\beta = c_5\alpha$ with $c_5 = [1 + 4{(1 + \gamma )^2}R_\pi ^2 + 4(1 + \gamma ){R_\pi }{G_*} + 8G_*^2]\nicefrac{1}{4\sigma}$.
Let the minibatch size $T \ge \max\{ \nicefrac{\log c_0^{-1} \beta}{\log\rho}, \nicefrac{\log \beta}{2\log\gamma}\}$, the finite-time convergence rate of HB-A2C algorithm is 
\begin{equation}\label{eqaa:25}
\frac{1}{4 K}\!\!\sum_{k=0}^{K-1} [ \norm{\nabla J(v_k)}^2 \!+\! \ee{ \norm{\Delta_k}^2 } ] 
\!\le\! \oo{ \frac{1}{\sqrt K} } + \oo{ \frac{1}{K} }
\end{equation}
where $\oo{ \nicefrac{1}{\sqrt K} } = [{\cal L}_K - {\cal L}_0 + ({c_3}\eta _1^{ - 1} + 2{R_w}{G_*} + {R_h}){R_h}{c_5} + ({c_4}\eta _1^{ - 1} + R_g^2)c_5^2 + (0.5L + L_*^2)R_h^2]\nicefrac{1}{\sqrt K}$ and $\oo{ \nicefrac{1}{K} } = \nicefrac{{2(1 - {\eta _1}){R_w}{R_g}{c_5}}}{{{\eta _1}K}}$ with the Lyapunov function as ${\cal L}_k = J(v_k) - \frac{1}{2} \ee{ \norm{\Delta_{k}}^2 }$. 
\end{corollary}

\begin{IEEEproof}
Define the Lyapunov function as ${\cal L}_k = J(v_k) - \frac{1}{2} \ee{ \norm{\Delta_{k}}^2 }$.
Summing \eqref{eqaa:23} and \eqref{eqaa:24} and dividing both sides by $\alpha$ with 
$\alpha = \Theta(\nicefrac{1}{\sqrt K})$ and 
$\beta = [1 + 4{(1 + \gamma )^2}R_\pi ^2 + 4(1 + \gamma ){R_\pi }{G_*} + 8G_*^2]\nicefrac{\alpha }{{4\sigma }}$, we obtain \eqref{eqaa:25}.  
\end{IEEEproof}

%Corollary \ref{cor:01} characterizes the unified convergence of actor and critic recursions with respect to total number of frames $K$.
%Based on Corollary \ref{cor:01}, we observe that the proposed HB-A2C finds an $\epsilon$-approximate stationary point with $\oo{\epsilon^{-2}}$ iterations for  reinforcement learning tasks with Markovian noise. 
%
%In our proposed HB-A2C algorithm, the learning rates of actor and critic recursions have the same order. 
%Moreover, we observe from $\oo{ \nicefrac{1}{\sqrt K} }$ that convergence rate is essentially controlled by the optimality drift term. 
%Based on $\oo{ \nicefrac{1}{\sqrt K} }$, we also observe that increasing the momentum factor $\eta_1$ can trade error introduced by the initial actor and critic parameters for the error introduced by the biased gradient descent recursions. 
%Our convergence rate $\oo{ \nicefrac{1}{\sqrt K} } + \oo{ \nicefrac{1}{K} }$ is tighter than those in \cite{Wu2020, ChenDec.2023}. 
%When compareing with the finite-time results of the A2C algorithms in \cite{Wu2020, ChenDec.2023}, our error bounds in \eqref{eqaa:25} hold for all $K \ge 1$, whereas those of \cite{Wu2020, ChenDec.2023} become available only after a mixing-time of updates.

Corollary~\ref{cor:01} characterizes the unified convergence of the actor and critic recursions with respect to the total number of frames $K$. Based on Corollary~\ref{cor:01}, we observe that the proposed HB-A2C algorithm finds an $\epsilon$-approximate stationary point with $\mathcal{O}(\epsilon^{-2})$ iterations for reinforcement learning tasks with Markovian noise.

In our proposed HB-A2C algorithm, the learning rates of the actor and critic recursions are of the same order. Furthermore, we observe from the term $\mathcal{O}( \nicefrac{1}{\sqrt{K}} )$ that the convergence rate is essentially controlled by the optimality drift term. Additionally, based on $\mathcal{O}( \nicefrac{1}{\sqrt{K}} )$, we observe that increasing the momentum factor $\eta_1$ can trade off the error introduced by the initial actor and critic parameters for the error introduced by the biased gradient descent recursions. Our convergence rate of $\mathcal{O}( \nicefrac{1}{\sqrt{K}} ) + \mathcal{O}( \nicefrac{1}{K} )$ is tighter than those in \cite{Wu2020, ChenDec.2023}. Compared to the finite-time results of the A2C algorithms in \cite{Wu2020, ChenDec.2023}, our error bounds in \eqref{eqaa:25} hold for all $K \geq 1$, whereas those of \cite{Wu2020, ChenDec.2023} become available only after a mixing time of updates.

\appendices
\section{Proof of Lemma \ref{lemmas:01}}\label{les:01}
For the critic parameter, the upper bound of the stoschatic semi-gradient in \eqref{eqaa:06} is derived as 
\begin{equation}\label{lemma01:01}
	\begin{split}
		\norm{g(w_k; O_k)} 
		&\le \norm{\Phi_{k}}\norm{w_k} + \norm{b_k} \\
		&\le (1+\gamma^T) R_w  + c_1(\gamma)R_r := R_g
	\end{split}
\end{equation}
where $c_1(\gamma) = \frac{1 - \gamma^T}{1-\gamma}$. 

Based on \eqref{lemma01:01}, the full semi-gradient $\ee{g(w_k; \bar O_k)}$ is upper-bounded as $\norm{ \ee{g(w_k; \bar O_k)} } \le \ee{ \norm{g(w_k; \bar O_k)} } \le R_g$.

Based on the definition in \eqref{eqaa:04}, we obtain the upper bound of $h(v_k, w_k^*; o_{k,t})$ as
\begin{equation}\label{lemma01:02}
	\begin{split}
		\norm{h(v_k, w_k^*; o_{k,t})} \le R_\pi[R_r \!+\! (1+\gamma)R_w] := R_h.
	\end{split}
\end{equation}

Based on \eqref{lemma01:02}, we obtain the upper bounds for the policy gradient and its $T$-step estimation as $\norm{\nabla J(v_k, w_k^*; s_{k,0})} \le R_h$ and $\norm{H(v_k, w_k; O_k)} \le R_h$.

\section{Proof of Lemma \ref{lemmas:02}}\label{les:02}
Following the definition of $\zeta(v_k, w_k; O_k)$, the difference between $\zeta(v_k, w_k; O_k)$ and $\zeta(v_k, w_{k-1}; O_k)$ can be decomposed as 
\begin{subequations}\label{lemma02:02}
\begin{align}
&\norm{ \zeta(v_k, w_k; O_k) - \zeta(v_k, w_{k-1}; O_k) } \\
& \le \norm{\Phi_{k}(w_k - w_{k-1})} + \norm{\ee{\bar\Phi_k}(w_k - w_{k-1})} \\
& \le 4 \norm{ w_k - w_{k-1} } 
\end{align}
\end{subequations}
where $\norm{\Phi_{k}} \le 1+\gamma^T \le 2$ and $\norm{\ee{\bar\Phi_k}} \le 2$. 

The gradient bias $\zeta(v_k, w_{k-1}; O_k)$ can be decomposed as 
\begin{equation}\label{lemma02:03}
\begin{split}
& \zeta(v_k, w_{k-1}; O_k) \\
&= g(w_{k-1}; O_k) - g(w_{k-1}; \tilde O_k) \\
&\hspace{0.3 cm} + g(w_{k-1}; \tilde O_k) - g(w_{k-1}; \bar O_k') \\
&\hspace{0.3 cm} + g(w_{k-1}; \bar O_k') - \ee{g(w_{k-1}; \bar O_k')} \\
&\hspace{0.3 cm} + \ee{g(w_{k-1}; \bar O_k')} - \ee{g(w_{k-1}; \bar O_k)}
\end{split}
\end{equation}
where observations $\tilde O_k$ are sampled from the behavior $\pi_{k-1}$ starting from $s_{k-1, 0}$, and $\bar O_k'$ are sampled from the stationary distribution $\mu_{k-1}\otimes\pi_{k-1}\otimes\mathds{P}$. 

Taking the expectation of $g(w_{k-1}; O_k) - g(w_{k-1}; \tilde O_k)$ conditional on ${\cal F}_{k-1}$, we obtain
\begin{equation}\label{lemma02:04}
\begin{split}
&\norm{\ee{ g(w_{k-1}; O_k) - g(w_{k-1}; \tilde O_k) | {\cal F}_{k-1}}} \\
&\le R_{g} \normtv{ \pp{ O_k \in \cdot| {\cal F}_{k-1} } - \pp{ \tilde O_k \in \cdot| {\cal F}_{k-1} } }.
\end{split}
\end{equation}

Following the definition of MDP and recalling $s_{k-1,T} = s_{k,0}$, we can expand the conditional probabilities in \eqref{lemma02:04} as
\begin{subequations}\label{lemma02:05}
\begin{align}
&\pp{ O_k \in \cdot | {\cal F}_{k-1} } \nonumber\\
&= {\cal P}_{k-1, T}\otimes\underbracket{\pi_k\otimes \mathds{P}\otimes\ldots\otimes\pi_k\otimes \mathds{P}}_{T~times}
		\label{lemma02:05a}\\
&\pp{ \tilde O_k \in \cdot | {\cal F}_{k-1} } \nonumber\\
&= {\cal P}_{k-1, T}\otimes\underbracket{\pi_{k-1}\otimes \mathds{P}\otimes\ldots\otimes\pi_{k-1}\otimes \mathds{P}}_{T~times}. \label{lemma02:05b}
\end{align}
\end{subequations}

Based on \eqref{lemma02:05}, we can upper bound $\normtv{ \pp{ O_k \in \cdot| {\cal F}_{k-1} } - \pp{ \tilde O_k \in \cdot| {\cal F}_{k-1} } }$ as 
\begin{multline}\label{lemma02:06}
	\normtv{ \pp{ O_k \in \cdot| {\cal F}_{k-1} } - \pp{ \tilde O_k \in \cdot| {\cal F}_{k-1} } } \\
	\le T|\mathcal A|L_\pi R_h \alpha.
\end{multline}

Substituting \eqref{lemma02:06} into \eqref{lemma02:04}, we obtain
\begin{multline}\label{lemma02:07}
	\norm{\ee{ g(w_{k-1}; O_k) - g(w_{k-1}; \tilde O_k) | {\cal F}_{k-1}}} \\
	\le T|{\cal A}| L_\pi R_g R_h \alpha.
\end{multline}

Taking the expectation of $g(w_{k-1}; \tilde O_k) - g(w_{k-1}; \bar O_k')$ conditional on ${\cal F}_{k-1}$, we obtain
\begin{multline}\label{lemma02:08}
	\norm{\ee{ g(w_{k-1}; \tilde O_k) - g(w_{k-1}; \bar O_k') | {\cal F}_{k-1}}} \\
	\le R_{g} \normtv{ \pp{ \tilde O_k = \cdot| {\cal F}_{k-1} } - \pp{\bar O_k' = \cdot| {\cal F}_{k-1}} }.
\end{multline}

Since the sample trajectory $\bar O_k'$ is obtained from the stationary distribution $\mu_{k-1}\otimes\pi_{k-1}\otimes\mathds{P}$, we have
\begin{equation}\label{lemma02:09}
\begin{split}
&\pp{ \bar O_k' \in \cdot | {\cal F}_{k-1} } \\
&= \mu_{k-1}\otimes\underbracket{\pi_{k-1}\otimes \mathds{P}\otimes\ldots\otimes\pi_{k-1}\otimes \mathds{P}}_{T~times}. 
\end{split}
\end{equation}

When Assumption \ref{assum:02} holds, we obtain the following upper bound based on Lemma \ref{lemmas:01} as
\begin{equation}\label{lemma02:10}
	\norm{\ee{ g(w_{k-1}; \tilde O_k) - g(w_{k-1}; \bar O_k') | {\cal F}_{k-1}}} 
	\!\le\! c_0 \rho^T\! R_g. \!
\end{equation}

Based on \eqref{lemma02:09}, we also have the following conditional expectation 
\begin{equation}\label{lemma02:11}
	\ee{ g(w_{k-1}; \bar O_k') - \ee{g(w_{k-1}; \bar O_k')} | {\cal F}_{k-1} } = 0
\end{equation}

Since trajectory $\bar O_k$ is sampled from stationary distribution induced by $\pi_{k}$, we have 
\begin{equation}\label{lemma02:12}
	\pp{ \bar O_k \in \cdot | {\cal F}_{k-1} }
	= \mu_{k}\otimes\underbracket{\pi_{k}\otimes \mathds{P}\otimes\ldots\otimes\pi_{k}\otimes \mathds{P}}_{T~times}. 
\end{equation}

Based on \eqref{lemma02:09} and \eqref{lemma02:12}, the norm of $\ee{g(w_{k-1}; \bar O_k')} - \ee{g(w_{k-1}; \bar O_k)}$ can be upper-bounded as
\begin{multline}\label{lemma02:13}
\norm{ \ee{g(w_{k-1}; \bar O_k')} - \ee{g(w_{k-1}; \bar O_k)} } \\
\le  (c_2 + T)|{\cal A}|L_\pi R_g R_h \alpha.
\end{multline}

Substituting \eqref{lemma02:07}, \eqref{lemma02:10}, \eqref{lemma02:11}, and \eqref{lemma02:13} into \eqref{lemma02:03}, we obtain upper bound of the expectation of gradient bias $\zeta(v_k, w_{k-1}; O_k)$ conditional on ${\cal F}_{k-1}$ as 
\begin{multline}\label{lemma02:14}
\norm{\ee{ \zeta(v_k, w_{k-1}; O_k) | {\cal F}_{k-1}} } \\
\le  (c_2 + 2T)|{\cal A}|L_\pi R_g \norm{v_k - v_{k-1}} + R_g\beta
\end{multline}
where $T \ge \nicefrac{\log c_0^{-1} \beta}{\log\rho}$.

\section{Proof of Lemma \ref{lemmas:05}}\label{les:05}
Sampling the state as $\bar s_{k,t} \sim \mu_v$ and the action as $\bar a_{k,t} \sim \pi_v$.
For the given actor parameter $v$, there always exists a unique optimal critic parameter $w^*_v$ that satisfies $\bar\Phi_v w^*_v = \bar b_v$ with 
$\bar\Phi_v = \ee{ \bar\phi_{k,0}[\bar\phi_{k,0} - \gamma^T\bar\phi_{k,T}]^\dag  } \succeq \sigma I$ and 
$\bar b_v = \ee{ \bar\phi_{k,0}\sum_{t=0}^{T-1}\gamma^t \bar r_{k,t} }$. 

Based on $\bar\Phi_v w^*_v - \bar b_v = 0$, we obtain 
\begin{equation}\label{lemma04:01}
	\nabla\bar\Phi_v w^*_v + \bar\Phi_v \nabla w^*_v = \nabla\bar b_v
\end{equation}
where $\nabla\bar b_v = \ee{  \bar\phi_{k,0}\sum_{t=0}^{T-1}\gamma^t \bar r_{k,t} \nabla\log\pi_v(a_{k,t}|s_{k,t}) }$ and $\nabla\bar\Phi_v w^*_v = \ee{ \bar\phi_{k,0}[\bar\phi_{k,0} - \gamma^T\bar\phi_{k,T}]^\dag w^*_v \nabla\log\pi_v(a_{k,t}|s_{k,t}) }$.

Based on \eqref{lemma04:01}, we obtain the Jacobian matrix as $\nabla w^*_v = \Phi_v^{-1}[ \nabla b_v - \nabla\Phi_v w^*_v ]$. Let two optimal critic parameters $w^*_v$ and $w^*_{v'}$ satisfy $\bar\Phi_v w^*_v = \bar b_v$ and $\bar\Phi_{v'} w^*_{v'} = \bar b_{v'}$. 
In order to derive the Lipschitz continuity of $w^*_v$ and bound of $\nabla w^*_v$,  we have the following inequalities based on Lemma \ref{lemmas:04} as
\begin{subequations}\label{lemma04:02}
	\begin{align}
		&\norm{\bar b_v - \bar b_{v'}} \le (1+c_2)c_1(\gamma)  R_r |{\cal A}|L_\pi \norm{v - v'} \label{lemma04:01a}\\
		&\norm{\bar\Phi_v - \bar\Phi_{v'}} \le 2( 1 + \gamma^T ) c_2 |{\cal A}|L_\pi \norm{v - v'} \label{lemma04:01b}\\
		&\norm{\bar b_v} \le c_1(\gamma)R_r, \norm{\bar \Phi_v^{-1}} \le \sigma^{-1} \label{lemma04:01c} \\
%		&\norm{\nabla\bar b_v - \nabla\bar b_{v'}} \le (1+c_2) c_1(\gamma)R_r R_\pi  |{\cal A}|L_\pi \norm{v - v'} \label{lemma04:01d} \\
%		&\norm{\nabla\bar\Phi_vw^*_v - \nabla\bar\Phi_{v'}w^*_v} \nonumber\\
%		&\le (1+\gamma^{T})(1+c_2) R_\pi R_w |{\cal A}|L_\pi \norm{v - v'} \label{lemma04:01e} \\
		&\norm{\nabla\bar b_v} \le c_1(\gamma)  R_r R_\pi, \norm{\nabla\bar \Phi_v w^*_v } \le (1+\gamma^{T})R_\pi R_w. \label{lemma04:01f}
	\end{align}
\end{subequations}

Then, we derive the Lipschitz continuity of $w^*_v$ as 
\begin{subequations}\label{lemma04:03}
	\begin{align}
		& \norm{w^*_v - w^*_{v'}} \\
		&= \norm{ \bar\Phi_v^{-1}\bar b_v -  \bar\Phi_{v'}^{-1}\bar b_{v'}} \\
		&\le \norm{ \bar\Phi_v^{-1}\bar b_v -  \bar\Phi_{v}^{-1}\bar b_{v'} } + \norm{ \bar\Phi_{v}^{-1}\bar b_{v'} - \bar\Phi_{v'}^{-1}\bar b_{v'} } \\
		&\le \norm{\bar\Phi_v^{-1}}\norm{\bar b_v -\bar b_{v'}} 
		\!+\! \norm{\bar \Phi_{v}^{-1}}\norm{\bar\Phi_{v} \!-\! \bar\Phi_{v'} }\norm{\bar\Phi_{v'}^{-1}} \norm{\bar b_{v'}} \\
		&\le L_* \norm{ v - v' }
	\end{align}
\end{subequations} 
where $L_* := [ 1 + c_2 + 2 (1+\gamma^T) \sigma^{-1} c_2 ]\sigma^{-1} c_1(\gamma) R_r |{\cal A}| L_{\pi}$.

Based on \eqref{lemma04:02}, the bound of the Jacobian matrix $\nabla w^*_v$ is derived as 
\begin{equation}
\norm{ \Phi_v^{-1}[ \nabla b_v - \nabla\Phi_v w^*_v ] }
\!\le\!  \norm{ \Phi_v^{-1} } \norm{ \nabla b_v - \nabla\Phi_v w^*_v } \!\le\! G_*
\end{equation}
where $G_* := \frac{R_\pi}{\sigma}[ c_1(\gamma)R_r \!+\! (1 \!+\! \gamma^T)R_w ]$.

%The Jacobian matrix is obtained as $\nabla w^*_v = \Phi_v^{-1}[ \nabla b_v - \nabla\Phi_v w^*_v ]$. 
%Based on the inequalities in \eqref{lemma04:02}, we obtain .

%Based on the inequalities in \eqref{lemma04:02}, we derive the Lipschitz continuity of $\nabla w^*_v$ as 
%\begin{subequations}\label{apdxb:03}
%\begin{align}
%&\norm{ \nabla w^*_v - \nabla w^*_{v'} } \\
%&= \| \Phi_v^{\!-\!1}[ \nabla b_v \!\!-\!\! \nabla\Phi_v w^*_v ] \!-\! \Phi_{v'}^{\!-\!1}[ \nabla b_{v'} - \nabla\Phi_{v'} w^*_{v'} ] \|  \\
%&\le \norm{ [\Phi_v^{-1} - \Phi_{v'}^{-1}] [ \nabla b_v - \nabla\Phi_v w^*_v ]  } \nonumber\\
%&\hspace{0.3 cm} + \norm{\Phi_{v'}^{-1}} \norm{\nabla b_v \!-\! \nabla b_{v'} \!+\! \nabla\Phi_{v'} w^*_{v'} - \nabla\Phi_{v'}w^*_v }\\
%&\le G_*\norm{v - v'}
%\end{align}
%\end{subequations}
%where the value of $G^*$ is obtained as $G_* =  [2\sigma^{-1}(1+\gamma^T)c_2(c_1(\gamma)R_r R_\pi + (1+\gamma^T)R_w R_\pi) + (1+c_2)c_1(\gamma)R_r R_\pi + (1+\gamma^T)(1+c_2)R_w R_\pi]\sigma^{-1}|{\cal A}| L_\pi $.

\section{Proof of Lemma \ref{lemmas:03}}\label{les:03}
Based on \eqref{eqaa:04}, we obtain
\begin{equation}\label{lemma03:01}
	\begin{split}
		\norm{h(v, w; o_{k,t}) \!-\! h(v, w'; o_{k,t})} 
		%&= \norm{[\gamma\phi_{k, t+1} - \phi_{k,t}]^{\dag}w\nabla\log\pi_v(a_{k,t} | s_{k,t}) - [\gamma\phi_{k, t+1} - \phi_{k,t}]^{\dag}w'\nabla\log\pi_v(a_{k,t} | s_{k,t}) } \\
		\le (1 \!+\! \gamma)R_\pi\norm{w - w'}.
	\end{split}
\end{equation}
Following the defintions in \eqref{eqaa:04} and \eqref{lemma03:01}, we obtain
\begin{equation}\label{lemma03:02}
	\norm{H(v, w; O_k) - H(v, w'; O_k)} 
	%	= \NORM{ (1-\gamma)\sum_{t=0}^{T-1} \gamma^t [h(v, w; o_{k,t}) - h(v, w'; o_{k,t})]}
	\le (1 + \gamma)R_\pi \norm{w - w'}
\end{equation}
where $w \in \mathbb{R}^{d_w}$ and $w' \in \mathbb{R}^{d_w}$.

\section{Proof of Theorem \ref{thm:02}}\label{sec:vi}
Before establishing the convergence of HB-A2C~actor, we first introduce several auxiliary inequlities.
Based on the policy gradient $\nabla J(v_k) = \ee{ \nabla J(v_k, w_k^*;s_{k,0}) }$ with $\nabla J(v_k, w_k^*;s_{k,0}) = (1-\gamma)\sum_{t=0}^{\infty} \gamma^t \int_{o_{k,t}} {\cal P}_{k,t}\otimes\pi_k\otimes\mathds{P}(o_{k,t}) h(v_k, w_k^*; o_{k,t}) d o_{k,t}$ and the stochastic policy gradient $H(v_k,w_k^*; O_k) =  (1-\gamma)\sum_{t=0}^{T-1} \gamma^t h(v_k, w_k^*; o_{k,t})$, we have 
\begin{align}
	&\nabla J(v_k, w_k^*;s_{k,0}) - \ee{H(v_k,w_k^*; O_k)|{\cal F}_{k}} \label{secvi:01}\\
	&= (1 \!-\! \gamma)\!\!\sum_{t=T}^{\infty} \gamma^t\!\! \int_{o_{k,t}}\!\!\!\! 
	{\cal P}_{k,t}\!\otimes\!\pi_k\!\otimes\!\mathds{P}(o_{k,t}) h(v_k, w_k^*; o_{k,t}) d o_{k,t}. \nonumber
\end{align}

Based on \eqref{secvi:01} and setting $T \ge \nicefrac{\log\beta}{2\log\gamma}$, we obtain
\begin{equation}\label{secvi:02}
	\begin{split}
		&\norm{ \nabla J(v_k) - \ee{H(v_k,w_k^*; O_k)}  }^2 \\
		&=\norm{ \ee{ \nabla J(v_k, w_k^*;s_{k,0}) - \ee{ H(v_k, w_k^*; O_k) | {\cal F}_k} }  }^2 \\
		&\le R_h^2\beta. 
	\end{split}
\end{equation}

Leveraging \eqref{secvi:02}, we obtain
\begin{subequations}\label{secvi:03}
	\begin{align}
		&\norm{ \nabla J(v_k) - \ee{H(v_k,w_k; O_k)}  }^2 \\
		&\le 2\norm{\nabla J(v_k) - \ee{H(v_k, w_k^*; O_k)} }^2 \nonumber\\
		&\hspace{0.2cm} + 2\norm{ \ee{H(v_k, w_k^*; O_k)} - \ee{H(v_k, w_k; O_k)} }^2 \\
		&\le 2R_h^2\beta + 2(1+\gamma)^2 R_\pi^2 \ee{ \norm{\Delta_k}^2 }  \label{secvi:03c}
	\end{align}
\end{subequations}
where \eqref{secvi:03c} follows from Lemma \ref{lemmas:03} and \eqref{secvi:02}.

Following the fact that $\inp{a, b} \ge \frac{1}{2}\norm{a}^2 - \frac{1}{2}\norm{a - b}^2$, we obtain the lower bound of  $\inp{ \nabla J(v_k), \ee{H(v_k, w_k; O_k)} }$ as 
\begin{equation}\label{secvi:04}
	\begin{split}
		&\inp{ \nabla J(v_k), \ee{H(v_k, w_k; O_k)} } \\
		&\ge \frac{1}{2}\norm{ \nabla J(v_k) }^2 - \frac{1}{2}\norm{ \nabla J(v_k) - \ee{H(v_k, w_k; O_k)} }^2\\
		&\ge \frac{1}{2}\norm{\nabla J(v_k)}^2 
		-  R_h^2 \beta 
		- (1 + \gamma)^2R_\pi^2 \ee{\norm{\Delta_k}^2 }.
	\end{split}
\end{equation}

Based on the Lipschitz continuity of the overall reward \eqref{eqaa:11} and the recursion in \eqref{eqaa:09}, we have 
\begin{equation}\label{secvi:10}
	\alpha \ee{\Omega_k } \le J(v_{k + 1}) - J(v_k) + \frac{1}{2} L R_h^2  \alpha^2
\end{equation}
where $\Omega_k = \inp{ \nabla J(v_k), H(v_k, w_k; O_k) }$

Summing \eqref{secvi:10} over $k = 0, 1, \ldots, K-1$, we have
\begin{equation}\label{secvi:11}
	\alpha \sum_{k=0}^{K-1}\ee{\Omega_k } \le J(v_K) - J(v_0) + \frac{1}{2} L R_h^2 \alpha^2 K.
\end{equation}

Substituting \eqref{secvi:11} into \eqref{secvi:04}, we have 
\begin{equation}\label{secvi:12}
	\begin{split}
		&\alpha\sum_{k=0}^{K-1}\!\Big[\frac{1}{2}\norm{\nabla J(v_k)}^2 \!-\! (1+\gamma)^2 R_\pi^2 \ee{ \norm{\Delta_k}^2 } \Big] \\
		&\le J(v_K) - J(v_0)
		+ \frac{1}{2} L R_h^2  \alpha^2 K  + R_h^2 \alpha\beta K. 
	\end{split}
\end{equation}

Dividing both sides of \eqref{secvi:12} by $K$, we complete the proof.

\section{Proof of Theorem \ref{thm:01}}\label{sec:v}
Based on the recursion \eqref{eqaa:07a}, we have $n_k = \eta_1\sum_{\tau=0}^k (1 - \eta_1)^{k - \tau} g(w_\tau; O_\tau)$ when  $n_{-1} = 0$.
Define $\Delta_k = w_{k} - w_k^*$, we start to analyze the convergence of the critic parameter $w_k$ by considering the following one-step drift 
\begin{equation}\label{secv:01}
	\begin{split}
		&\frac{1}{2}\norm{\Delta_{k+1}} - \frac{1}{2}\norm{\Delta_k}^2 \\
		&\le \frac{1}{2}\norm{ \beta n_k +  w_{k+1}^* - w_{k}^* }^2 \hspace{0.2 cm}\mbox{(gradient variance)}  \\
		&\hspace{0.3 cm} + \inp{ \Delta_k, w_k^* - w_{k + 1}^* } \hspace{0.84 cm}\mbox{(optimality drift)}\\
		&\hspace{0.3 cm} - \beta \inp{\Delta_k, n_k}  \hspace{1.88 cm}\mbox{(gradient progress)}
	\end{split}
\end{equation}
where $n_k$ is obtained via recursion \eqref{eqaa:07b}. 

\subsection{Analysis of the Gradient Variance}
We start by analyzing the gradient variance term as 
\begin{equation}\label{secv:02}
	\frac{1}{2}\norm{ \beta n_k +  w_{k+1}^* - w_{k}^* }^2
	\!\le\! \beta^2\norm{n_k}^2 + \norm{w_{k+1}^* - w_k^*}^2 \!\!
\end{equation}
where the inequality follows $\frac{1}{2}\norm{a + b}^2 \le \norm{a}^2 + \norm{b}^2$.

Based on \eqref{eqaa:16}, the first term on the right-hand side of \eqref{secv:02} is upper-bounded as 
\begin{equation}\label{secv:03}
	\beta^2\norm{n_k}^2 \le R_g^2 \beta^2.
\end{equation}

Based on Lemma \ref{lemmas:05}, the second term on the right-hand side of \eqref{secv:02} is upper-bounded as
\begin{equation}\label{secv:04}
	\norm{ w_k^* - w_{k+1}^* }^2
	\le  L_*^2 \norm{v_{k+1} - v_k}^2 \le L_*^2 R_h^2 \alpha^2.
\end{equation}

Summing \eqref{secv:03} and \eqref{secv:04}, we obtain the upper bound of the gradient variance term as 
\begin{equation}\label{secv:05}
	\frac{1}{2}\norm{ \beta n_k +  w_{k+1}^* \!-\! w_{k}^* }^2 \le R_g^2 \beta^2 + L_*^2 R_h^2 \alpha^2.
\end{equation}

\subsection{Analysis of the Optimality Drift}
Based on Lemma \ref{lemmas:05}, there exits a Jacobian matrix $\nabla^{\dag} w^*_v$ such that $w^*_k \!-\!  w^*_{k+1} = \nabla^{\dag} w^*_v(v_k - v_{k+1})$. 
Therefore, we can recast the optimality drift in \eqref{secv:01} as 
\begin{subequations}\label{secv:06}
\begin{align}
& \inp{ \Delta_k,  w^*_k -  w^*_{k+1} } \\
&= \inp{ \nabla w^*_v \Delta_k, v_k - v_{k+1} } \\
&= -\alpha \inp{ \nabla w^*_v \Delta_k, H(v_k, w_k; O_k) }  \label{secv:06b} \\
&= -\alpha \inp{ \nabla w^*_v \Delta_k, H(v_k, w_k; O_k) - H(v_k, w_k^*; O_k) } \nonumber\\
&\hspace{0.3cm} - \alpha \inp{ \nabla w^*_v \Delta_k, H(v_k, w_k^*; O_k) - \nabla J(v_k) } \nonumber\\
&\hspace{0.3cm} - \alpha \inp{ \nabla w^*_v \Delta_k, \nabla J(v_k) }  \label{secv:06c}
\end{align}
\end{subequations}
where \eqref{secv:06b} follows from the recursion in \eqref{eqaa:09}.

Based on Lemma \ref{lemmas:03}, the three terms on the right-hand side of \eqref{secv:06c} can be bounded as 
\begin{subequations}\label{secv:006}
\begin{align}
&| \inp{ \nabla w^*_v \Delta_k, H(v_k, w_k; O_k) - H(v_k, w_k^*; O_k) } | \nonumber\\
&\le (1+\gamma)R_\pi G_* \norm{\Delta_k}^2\\
&| \ee{\inp{ \nabla w^*_v \Delta_k, H(v_k, w_k^*; O_k) - \nabla J(v_k) }| {\cal F}_k} | \nonumber\\
&\le 2\beta  R_w R_h G_* \\
&|\inp{ \nabla w^*_v \Delta_k, \nabla J(v_k) }  |
\le  2 G_*^2 \norm{\Delta_k}^2 + \frac{1}{4}\norm{\nabla J(v_k)}^2.
\end{align}
\end{subequations}

Substituting \eqref{secv:006} into the expectation of \eqref{secv:06c}, we have 
\begin{multline}\label{secv:0006}
\ee{ \inp{ \Delta_k,  w^*_k -  w^*_{k+1} } } 
\le [(1+\gamma)R_\pi G_* + 2G_*^2]\alpha\ee{ \norm{\Delta_k}^2 } \\
+ \frac{1}{4}\alpha\norm{\nabla J(v_k)}^2 + 2\alpha \beta  R_w R_h G_*. 
\end{multline}

Substituting \eqref{secv:05} and \eqref{secv:0006} into the expectation of \eqref{secv:01}, we have
\begin{equation}\label{secv:17}
\begin{split}
&\beta\ee{ \inp{\Delta_k, n_k} } -  [(1+\gamma)R_\pi G_* + 2G_*^2]\alpha \ee{\norm{\Delta_k}^2} \\
&\le \frac{1}{2}[\ee{\norm{\Delta_k}^2} - \ee{\norm{\Delta_{k+1}}^2}] 
+ \frac{1}{4}\alpha\norm{\nabla J(v_k)}^2  \\
&\hspace{0.3cm}+ 2\alpha\beta R_w R_h G_* 
\!+\! R_g^2 \beta^2 
\!+\! L_*^2 R_h^2 \alpha^2. \!
\end{split}
\end{equation}

\subsection{Analysis of the Gradient Progress}
Based on \eqref{eqaa:07a}, we can decompose the gradient progress term in \eqref{secv:01} as 
\begin{subequations}\label{secv:07}
\begin{align}
& \inp{\Delta_{k}, n_k} \\
&= \eta_1\inp{\Delta_k, g(w_k; O_k)} + (1-\eta_1)\inp{\Delta_k, n_{k-1}} \\
&= \eta_1\inp{\Delta_k, g(w_k; O_k)} + (1-\eta_1)\inp{\Delta_{k - 1}, n_{k-1}} \nonumber \\
&\hspace{0.2 cm} + (1-\eta_1)\inp{\Delta_k - \Delta_{k-1}, n_{k-1}}. 
\end{align}
\end{subequations}

The second term in \eqref{secv:07} is lower-bounded as
\begin{subequations}\label{secv:08}
	\begin{align}
		&(1-\eta_1)\inp{\Delta_k - \Delta_{k-1}, n_{k-1}} \\
		&= (1-\eta_1)\inp{w_k - w_{k-1} + w_{k-1}^* - w_k^*, n_{k-1}} \\
		&\ge -(1-\eta_1)\norm{n_{k-1}}[ \norm{w_k - w_{k-1}} + \norm{w^*_k - w^*_{k-1}} ] \\
		&\ge -(1-\eta_1) R_g [R_g \beta + L_*\norm{v_k - v_{k-1}}] \label{secv:08c}
	\end{align}
\end{subequations}
where \eqref{secv:08c} follows from  \eqref{eqaa:16}, \eqref{eqaa:17}, and Lemma \ref{lemmas:05}.

Substituting the expectation of \eqref{secv:08} into the expectation of \eqref{secv:07}, we obtain
\begin{equation}\label{secv:09}
\begin{split}
&\eta_1 \ee{ \inp{\Delta_k, g(w_k; O_k)} } \\
&\le (1-\eta_1) R_g [R_g \beta + L_*\ee{ \norm{v_k - v_{k-1}} }]\\
&\hspace{0.3 cm} + \ee{ \inp{\Delta_{k}, n_k}  } - (1-\eta_1)\ee{ \inp{\Delta_{k - 1}, n_{k-1}}  }.
\end{split}
\end{equation}

Based on \eqref{eqaa:06}, the left-hand side of \eqref{secv:09} can be recast as 
\begin{subequations}\label{secv:10}
	\begin{align}
		&\eta_1\inp{\Delta_k, g(w_k; O_k)} \\
		&= \eta_1\inp{\Delta_k, \ee{g(w_k; \bar O_k)} }
		+ \eta_1\inp{\Delta_k, \zeta(v_k, w_k; O_k)} \\
		&= \eta_1\inp{\Delta_k, \ee{g(w_k; \bar O_k)} - \ee{g(w_k^*; \bar O_k)} } \nonumber\\
		&\hspace{0.3cm} + \eta_1\inp{\Delta_k, \zeta(v_k, w_k; O_k)} \label{secv:09b} \\
		&\ge \eta_1\sigma \norm{\Delta_k^*}^2
		+ \eta_1\inp{\Delta_k, \zeta(v_k, w_k; O_k)} \label{secv:09c} \\
		&\ge \eta_1\sigma \norm{\Delta_k^*}^2 
		 + \eta_1\inp{\Delta_k - \Delta_{k-1}, \zeta(v_k, w_k; O_k)} \nonumber\\
		&\hspace{0.3cm} + \eta_1\inp{\Delta_{k-1}, \zeta(v_k, w_k; O_k) - \zeta(v_k, w_{k-1}; O_k)} \nonumber\\
		&\hspace{0.3cm} + \eta_1\inp{\Delta_{k-1}, \zeta(v_k, w_{k-1}; O_k)} \label{secv:09d}
	\end{align}
\end{subequations}
where \eqref{secv:09b} is based on the fact $\ee{g(w_k^*; \bar O_k)} = 0$, 
and \eqref{secv:09c} is based on \eqref{eqaa:14}.

Based on \eqref{eqaa:06} and Lemma \ref{lemmas:01}, we obtain the upper bound of $\zeta(v_k, w_k; O_k)$ as $\norm{\zeta(v_k, w_k; O_k)} \le 2 R_g$. 
Together with the inequality \eqref{eqaa:17} and the Lipschitz continuity in Lemma \ref{lemmas:05}, 
we derive the upper bounds for the three terms on the right-hand side of \eqref{secv:09d} as 
\begin{subequations}\label{secv:11}
	\begin{align}
		&| \inp{\Delta_{k} - \Delta_{k-1}, \zeta(v_k, w_k; O_k)} | \nonumber\\
		&\le 2R_g\big[R_g\beta + L_*\norm{v_k - v_{k-1}} \big] \label{secv:11a} \\
		&| \inp{\Delta_{k-1}, \zeta(v_k, w_k; O_k) - \zeta(v_k, w_{k-1}; O_k)} | \nonumber\\
		&\le 16R_w R_g \beta \label{secv:11b}\\
		&| \ee{\inp{\Delta_{k-1}, \zeta(v_k, w_{k-1}; O_k)} |{\cal F}_{k-1}} | \nonumber\\
		&\le 2R_w R_g [\beta + (c_2 + 2T)|{\cal A}|L_\pi \norm{v_k - v_{k-1}} ] \label{secv:11c}
	\end{align}
\end{subequations}
with $T \ge \nicefrac{\log c_0^{-1} \beta}{\log\rho}$.

Summing \eqref{secv:11a}--\eqref{secv:11c}, we obtain
\begin{equation}\label{secv:12}
\begin{split}
	&| \ee{\inp{\Delta_k, \zeta(v_k, w_k; O_k)} | {\cal F}_{k-1}} | \\
	&\le c_3' \ee{\norm{v_k - v_{k-1}} | {\cal F}_{k-1}} + c_4' \beta
\end{split}
\end{equation}
where $c_3' := 2R_g [ L_* + (c_2 + 2T)|{\cal A}|L_\pi R_w ]$ and $c_4' := 2R_g(R_g + 9 R_w)$. 

Substituting \eqref{secv:12} into \eqref{secv:10} and taking iterated expectation, we obtain
\begin{equation}\label{secv:13}
\begin{split}
&\eta_1 \ee{ \inp{\Delta_k, g(w_k; O_k)} } \\
&\ge \eta_1 \sigma \ee{ \norm{\Delta_k}^2 } 
- \eta_1 c_3'\ee{\norm{v_k - v_{k-1}}}  
- \eta_1 c_4'\beta. 
\end{split}
\end{equation}

Substituting \eqref{secv:13} into \eqref{secv:09}, we obtain
\begin{equation}\label{secv:14}
\begin{split}
\eta_1 \sigma \ee{ \norm{\Delta_k}^2 } 
&\!\le\! c_3 \ee{ \norm{v_k - v_{k-1}} }
\!+\! c_4 \beta \!+\! \ee{ \inp{\Delta_{k}, n_k}  } \\
&\hspace{0.3cm}  - (1-\eta_1)\ee{ \inp{\Delta_{k - 1}, n_{k-1}}  }
\end{split}
\end{equation}
where $c_3 := [(1+\eta_1)L_* + 2\eta_1(c_2 + 2T)|{\cal A}|L_\pi R_w]R_g$ and $c_4 := [2\eta_1(R_g + 9R_w) + (1-\eta_1)R_g]R_g$.

Summing \eqref{secv:14} over $k = 0, \ldots, K-1$ and recalling the fact $n_{-1} = 0$, we obtain
\begin{subequations}\label{secv:15}
\begin{align}
&\eta_1 \sigma \sum_{k=0}^{K-1}\ee{ \norm{\Delta_k}^2 } \\
&\le c_3 \sum_{k=0}^{K-1} \ee{ \norm{v_k - v_{k-1}} }
	+ c_4 \beta K 
	+ \eta_1 \sum_{k=0}^{K-1}\ee{ \inp{\Delta_{k}, n_{k}}  } \nonumber\\
&\hspace{0.3cm} + (1-\eta_1)\ee{ \inp{\Delta_{K-1}, n_{K-1}}  } 
	 \label{secv:15a}\\
&\le c_3 \sum_{k=0}^{K-1} \ee{ \norm{v_k - v_{k-1}} }
	+ c_4 \beta K \nonumber\\
&\hspace{0.3cm} + 2 (1-\eta_1) R_w R_g
	+ \eta_1 \sum_{k=0}^{K-1}\ee{ \inp{\Delta_{k}, n_{k}}  } \label{secv:15b}
\end{align}
\end{subequations}
where \eqref{secv:15b} follows from the facts $\norm{\Delta_{K-1}} \le 2R_w$ and $\norm{n_{K-1}} \le R_g$.

Multipying both sides of \eqref{secv:15} by $\nicefrac{\beta}{\eta_1}$, we have 
\begin{subequations}\label{secv:16}
\begin{align}
&\sigma \beta\sum_{k=0}^{K-1}\ee{ \norm{\Delta_k}^2 } \\
&\le \frac{c_3 \beta}{\eta_1} \sum_{k=0}^{K-1} \ee{ \norm{v_k - v_{k-1}} } 
 + \frac{c_4}{\eta_1} \beta^2 K  \nonumber\\
&\hspace{0.3cm} + \frac{2(1-\eta_1)}{\eta_1} R_w R_g \beta 
 + \beta\sum_{k=0}^{K-1}\ee{ \inp{\Delta_{k}, n_{k}}  } \\
&\le \frac{c_3}{\eta_1} R_h \alpha\beta K 
		+ \frac{c_4}{\eta_1} \beta^2 K
		+ \frac{2(1-\eta_1)}{\eta_1} R_w R_g \beta  \nonumber\\
&\hspace{0.3cm} + \beta\sum_{k=0}^{K-1}\ee{ \inp{\Delta_{k}, n_{k}}  }  \label{secv:16b}
\end{align}
\end{subequations}
where \eqref{secv:16b} follows from the inequality in \eqref{eqaa:19}.

Substituting \eqref{secv:17} into \eqref{secv:16} and performing several algebraic manipulations, we obtain
\begin{equation}\label{secv:18}
\begin{split}
&\Big[\sigma \beta - [(1+\gamma)R_\pi G_* + 2G_*^2]\alpha\Big]\sum_{k=0}^{K-1}\ee{ \norm{\Delta_k}^2 }  \\
&\le \frac{\alpha}{4}\sum_{k=0}^{K-1}\norm{\nabla J(v_k)}^2
+ \frac{1}{2}[\ee{\norm{\Delta_0}^2} - \ee{\norm{\Delta_K}^2}] \\
&\hspace{0.3cm} + L_*^2 R_h^2 \alpha^2 K
+ [\frac{c_4}{\eta_1} + R_g^2 ]\beta^2 K \\
&\hspace{0.3cm} + [\frac{c_3}{\eta_1} + 2 R_w G_*] R_h \alpha\beta K 
+ \frac{2(1-\eta_1)}{\eta_1} R_w R_g \beta.
\end{split}
\end{equation}

Dividing both sides of \eqref{secv:18} by $K$, we complete the proof.

\section{Supporting Lemmas}\label{apdx:supportlemma}
\begin{lemma}\label{lemmas:04}
	When Assumption \ref{assum:01} is satisfied, the joint distribution satisfies 
	\begin{equation}\label{lemma05:00}
		\normtv{ \mu_v\otimes\pi_v - \mu_{v'}\otimes\pi_{v'} } \le (1+c_2)|{\cal A}|L_\pi \norm{v - v'}
	\end{equation}
	where $c_2 > 0$, $v \in \mathbb{R}^{d_v}$, and  $v' \in \mathbb{R}^{d_v}$.
\end{lemma}

\begin{IEEEproof}
	Following \cite[Corollary 3.1]{Mitrophanov2005}, we have
	\begin{subequations}\label{lemma05:01}
		\begin{align}
			&\normtv{\mu_v - \mu_{v'}} \\
			&\le c_2\NORMTV{\sum_a\pi_v\otimes\mathds{P} - \sum_a \pi_{v'}\otimes\mathds{P}} \label{lemma05:01a}\\
			&\le c_2\int_{a, s'} \pp{ s'|s,a } |\pi_v(a|s) - \pi_{v'}(a|s)| da ds'  \label{lemma05:01b}\\
			&\le c_2|{\cal A}|L_\pi\norm{v - v'} \label{lemma05:01c}
		\end{align}
	\end{subequations}
	where $c_2$ is a predetermined positive constant, and \eqref{lemma05:01c} follows the $L_\pi$-Lipschitz of behavior policy.

Then, we start to analyze the total variation norm for $\mu_v\otimes\pi_v - \mu_{v'}\otimes\pi_{v'}$ as
\begin{subequations}\label{lemma05:02}
\begin{align}
&\normtv{  \mu_v\otimes\pi_v - \mu_{v'}\otimes\pi_{v'} } \\
&= \int_{s,a}  | \mu_v(s)\pi_v(a|s) - \mu_{v'}(s)\pi_{v'}(a|s) | ds da \\
&\le \int_{s,a} \mu_v(s)| \pi_v(a|s) - \pi_{v'}(a|s) | ds da \nonumber\\
&\hspace{0.3 cm} + \int_{s,a} \pi_{v'}(a|s) | \mu_{v}(s) - \mu_{v'}(s) | ds da \\
&\le (1+c_2)|{\cal A}|L_\pi \norm{v - v'} \label{lemma05:02d}
\end{align}
\end{subequations}
where \eqref{lemma05:02d} follows from \eqref{lemma05:01} and $L_\pi$-Lipschitz continuity of behavior. 
\end{IEEEproof}

\begin{lemma}\label{lemmas:06}
	When Assumption \ref{assum:01} holds, the overall reward $\nabla J(v)$ has $L$-Lipschitz continuous gradient. 
\end{lemma}
\begin{IEEEproof}
	Based on Lemma \ref{lemmas:05}, the optimal critic $w^*(v)$ is $L_*$-Lipschitz with respect to $v$. 
	Together with the similar arguments in \cite[Lemma 3.2]{Zhang2020}, there exits a positive constant $L$ such that $\norm{\nabla J(v) - \nabla J(v')} \le L\norm{v - v'}$. Therefore, we can apply the equivalent condition to $\norm{\nabla J(v) - \nabla J(v')} \le L\norm{v - v'}$ in \cite[Theorem 2.1.5]{Nesterov2018} in order to obtain \eqref{eqaa:11}. 
\end{IEEEproof}

%\balance
\bibliographystyle{IEEEtran}
\bibliography{new_RL}

\begin{IEEEbiographynophoto}{Yanjie Dong (Member, IEEE)} is an Associate Professor and the Assistant Dean of Artificial Intelligence Research Institute, Shenzhen MSU-BIT University. 
	Dr. Dong respectively obtained his Ph.D. and M.A.Sc. degree from The University of British Columbia, Canada, in 2020 and 2016. 
	His research interests focus on the design and analysis of machine learning algorithms, machine learning based resource allocation algorithms, and quantum computing technologies. 
%	He regularly serves as a member of Technical Program Committee in flagship conferences in IEEE ComSoc.
\end{IEEEbiographynophoto}

\vspace{-0.5 cm}

\begin{IEEEbiographynophoto}{Haijun Zhang (Fellow, IEEE)}
	is a Professor at the University of Science and Technology Beijing, China. He was a postdoctoral research fellow in the Department of Electrical and Computer Engineering at The University of British Columbia, Canada. 
%	He serves/served as Track Co-Chair of WCNC 2020, Symposium Chair of Globecom’19, TPC Co-Chair of INFOCOM 2018 Workshop on Integrating Edge Computing, Caching, and Offloading in Next Generation Networks, and General Co-Chair of GameNets’16. 
	He serves/served as an Editor of IEEE Transactions on Information Forensics and Security, IEEE Transactions on Communications, IEEE Transactions on Network Science and Engineering, and IEEE Transactions on Vehicular Technology. He received the IEEE CSIM Technical Committee Best Journal Paper Award, in 2018, IEEE ComSoc Young Author Best Paper Award, in 2017, and IEEE ComSoc Asia-Pacific Best Young Researcher Award, in 2019. He is an IEEE ComSoc Distinguished Lecturer.
\end{IEEEbiographynophoto}

\vspace{-0.5 cm}

\begin{IEEEbiographynophoto}{Gang Wang (Senior Member, IEEE)} is a Professor with the School of Automation at the Beijing Institute of Technology.
	Dr. Wang received a B.Eng. degree in 2011, and a Ph.D. degree  in 2018, both from the Beijing Institute of Technology, Beijing, China. 
	He also hold a Ph.D. degree   from the University of Minnesota, Minneapolis, USA, in 2018, where he stayed as a postdoctoral researcher until July 2020. 
	His research interests focus on the areas of signal processing, control and reinforcement learning with applications to cyber-physical systems and multi-agent systems. 
	He was the recipient of the Best Paper Award from the Frontiers of Information Technology \& Electronic Engineering  in 2021, the Excellent Doctoral Dissertation Award from the Chinese Association of Automation in 2019, the outstanding editorial board member award from the IEEE Signal Processing Society in 2023.
	He serves as an Editor of Signal Processing and IEEE Transactions on Signal and Information Processing over Networks. 
\end{IEEEbiographynophoto}

\vspace{-0.5 cm}

\begin{IEEEbiographynophoto}{Shisheng Cui} is a Professor with the School of Automation at the Beijing Institute of Technology.
Dr. Cui received the B.S. degree from Tsinghua University, Beijing, China, in 2009, the M.S. degree  from Stanford University, Stanford, USA, in 2011 and the Ph.D. degree  from Pennsylvania State University, University Park, USA, in 2019. His current research interests lie in optimization, variational inequality problems, and inclusion problems complicated by nonsmoothness and uncertainty.
\end{IEEEbiographynophoto}

\vspace{-0.5 cm}

\begin{IEEEbiographynophoto}{Xiping Hu} 
	is currently a Professor with Shenzhen MSU-BIT University, and is also with Beijing Institute of Technology, China. Dr. Hu received the PhD degree from the University of British Columbia, Vancouver, BC, Canada. 
	Dr. Hu is the co-founder and chief scientist of Erudite Education Group Limited, Hong Kong, a leading language learning mobile application company with over 100 million users, and listed as top 2 language education platform globally. His research interests include affective computing, mobile cyber-physical systems, crowdsensing, social networks, and cloud computing. He has published more than 150 papers  in the prestigious conferences and journals, such as IJCAI, AAAI, ACM MobiCom, WWW, and IEEE TPAMI/TMM/TVT/IoTJ/COMMAG.
\end{IEEEbiographynophoto}

\end{document}